\documentclass{article} 
\usepackage{colm2024_conference}

\usepackage{microtype}
\usepackage{hyperref}
\usepackage{url}
\usepackage{booktabs}
\usepackage{graphicx}
\usepackage{setspace}
\usepackage{enumitem}
\usepackage{times}
\usepackage{latexsym}
\usepackage{booktabs}  

\usepackage[T1]{fontenc}

\usepackage[utf8]{inputenc}

\usepackage{microtype}

\usepackage{inconsolata}
\usepackage{fancyhdr}

\usepackage{graphicx}
\usepackage{subcaption}
\usepackage{multirow}  
\usepackage{tcolorbox}

\newtcolorbox{promptbox}[2][]{%
    colback=gray!5,
    colframe=gray!50,
    boxrule=0.5pt,
    arc=0mm,
    title=#2,
    fonttitle=\bfseries,
    #1
}

\title{Revisiting the Test-Time Scaling of o1-like Models: Do they Truly Possess Test-Time Scaling Capabilities? }

\colmfinalcopy

\author{Zhiyuan Zeng\textsuperscript{1},\quad Qinyuan Cheng\textsuperscript{1},\quad Zhangyue Yin\textsuperscript{1},\quad Yunhua Zhou\textsuperscript{2}, \quad Xipeng Qiu\textsuperscript{1}\thanks{~Corresponding author} \\
$^1$School of Computer Science, Fudan University, Shanghai, China \\
$^2$Shanghai AI Laboratory \\
{cengzy23@m.fudan.edu.cn; xpqiu@fudan.edu.cn}
}

\fancypagestyle{firstpage}{
  \lhead{\begin{picture}(0,0)\put(0,-6){\includegraphics[width=0.3\linewidth]{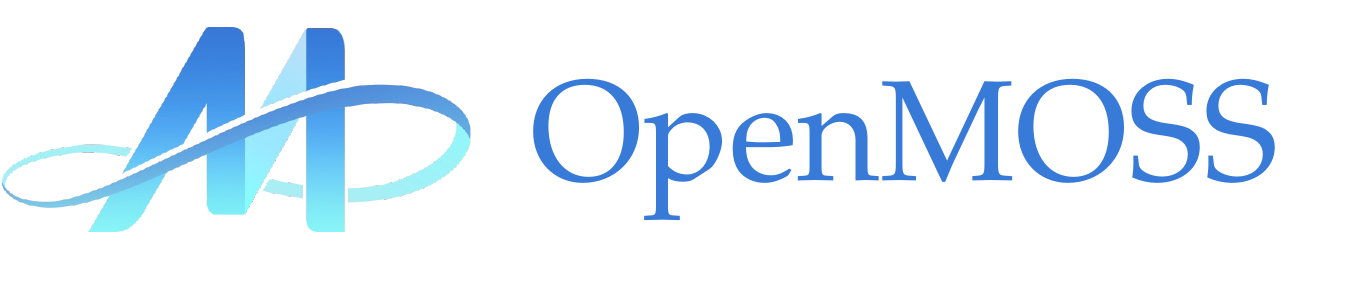}}\end{picture}}
}

\begin{document}
\thispagestyle{firstpage}

\maketitle
\begin{abstract}
The advent of test-time scaling in large language models (LLMs), exemplified by OpenAI’s o1 series, has advanced reasoning capabilities by scaling computational resource allocation during inference. While successors like QwQ, Deepseek-R1 (R1) and LIMO replicate these advancements, whether these models truly possess test-time scaling capabilities remains underexplored. This study found that longer CoTs of these o1-like models do not consistently enhance accuracy; in fact, correct solutions are often shorter than incorrect ones for the same questions. Further investigation shows this phenomenon is closely related to models' self-revision capabilities - longer CoTs contain more self-revisions, which often lead to performance degradation. We then compare sequential and parallel scaling strategies on QwQ, R1 and LIMO, finding that parallel scaling achieves better coverage and scalability. Based on these insights, we propose ``Shortest Majority Vote'', a method that combines parallel scaling strategies with CoT length characteristics, significantly improving models' test-time scalability compared to conventional majority voting approaches. Our implementation is available at \url{https://github.com/ZhiYuanZeng/test-time-scaling-eval}.


\end{abstract}
\section{Introduction}
The release of the OpenAI o1 series models~\citep{o1_blog,o1_system_card} marked a pivotal advancement in the reasoning capabilities of Large Language Models (LLMs), introducing a novel scaling paradigm, test-time scaling, which allocates more compute resources during test time. The test-time scaling have two dimensions, sequential and parallel \citep{o1-roadmap}. Sequential scaling increase test-time compute by scaling the length of Chain-of-Thought (CoT)~\citep{wei2022chain}, while parallel scaling parallely samples multiple solutions and pick the best one. 

Following o1's success, models such as QwQ \citep{qwq}, Deepseek-R1 (R1)~\citep{deepseek-r1} and LIMO \citep{ye2025limoreasoning} have emerged as leading open-source successors, replicating o1's achievements and demonstrating comparable reasoning abilities. Although both QwQ, R1 and LIMO demonstrate strong reasoning capabilities and the ability to generate lengthy CoT at test time, the existence of \textbf{true test-time scaling where performance consistently improves with longer CoTs} remains to be verified for these models.

\begin{figure}[t]
    \centering
    \includegraphics[width=1.0\linewidth]{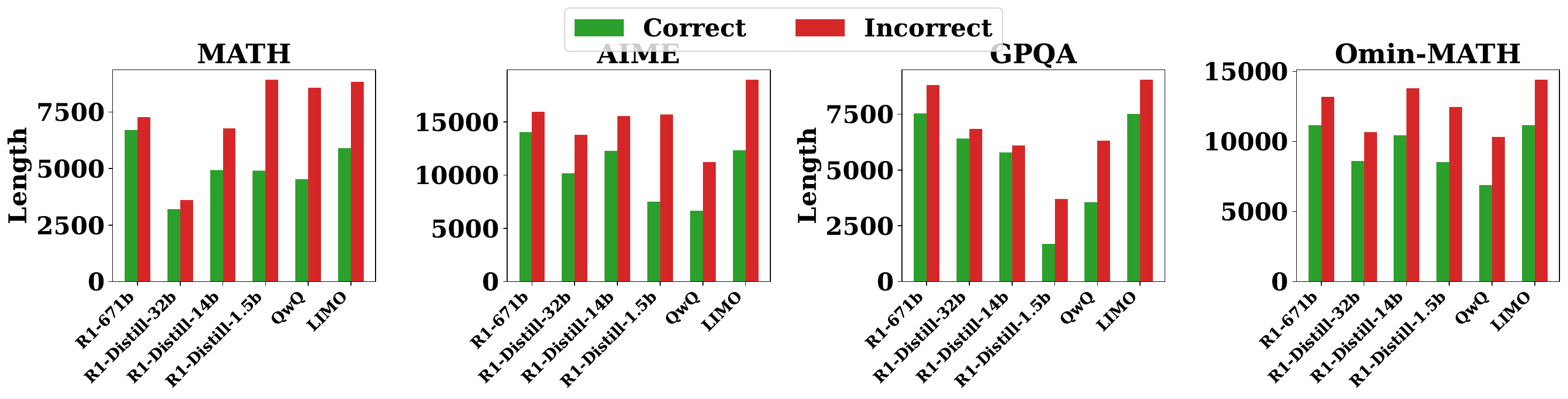}
    \caption{The average length of correct solutions versus incorrect solutions evaluated on the same questions. For each question, solution lengths were averaged separately for correct and incorrect responses, then averaged across all questions.}
    \label{fig:overall-correct-vs-incorrect}
\end{figure}

To explore this question, we systematically investigate the relationship between CoT length and reasoning performance in QwQ, R1 and LIMO, challenging the conventional assumption that extended reasoning chains inherently lead to improved accuracy. Contrary to expectations, our analysis reveals that longer CoTs do not consistently improve accuracy of these o1-like models.
Notably, we found that the average length of correct solutions is shorter than that of incorrect ones for the same questions, which is shown in Figure \ref{fig:overall-correct-vs-incorrect}. This counterintuitive finding underscores the need for a deeper understanding of the test-time scaling of o1-like models.

To understand why the longer CoTs do not lead to the better performance, we compared the difference between long CoTs and short CoTs, finding that long CoTs contain more self-revisions (``Wait'', ``Alternatively'') than the short CoTs, which is shown in Appendix \ref{app:revision-examples}. Inspired by that, we iteratively prompted QwQ, R1 and LIMO for more self-revisions. Our observations revealed that QwQ and R1-Distill-1.5b exhibited performance degradation as the length of reflection increased. In contrast, R1-Distill-14b, R1-Distill-32b, and LIMO demonstrated initial performance improvements during early revisions, followed by oscillatory behavior in subsequent iterations. To further understand the limitations of sequential scaling, we evaluated the models' capacity to revise incorrect answers. Our findings indicate that QwQ, R1 and LIMO all demonstrated limited ability to convert incorrect answers to correct ones during the revision process. Most revisions retained the original answers, and more concerning, both QwQ and R1-Distill-1.5b showed a higher propensity to change correct answers to incorrect ones rather than vice versa. These results reveal that \textbf{self-revision ability is a key factor in the effectiveness of sequential scaling for o1-like models}.

Given the limited effectiveness of sequential scaling, we explored an alternative test-time scaling strategie, parallel scaling. Our comparative analysis of sequential and parallel scaling revealed that parallel scaling not only achieves the better coverage (pass@k score) but also offers superior scalability compared to sequential scaling for QwQ and R1, which demonstrates that o1-like models have limited sequential-scaling capability, but strong parallel-scaling capability.

Building on these findings, we propose a novel test-time scaling method, \textbf{Shortest Majority Vote}, which incorporate parallel scaling approaches with our insight on sequential scaling. In particular, this method leverages the observation that shorter solutions tend to lead to better performance compared to longer ones. Shortest Majority Vote improves majority vote by prioritizing clusters that have both more solutions and shorter solution lengths. Experimental results demonstrate that Shortest Majority Vote substantially outperforms conventional Majority Vote, significantly improving the test-time scalability of both QwQ and R1 models.

Our contributions are as follows:
\begin{enumerate}[itemsep=0pt,parsep=0pt,label=\arabic*)]
    \item We systematically investigate the test-time scaling capabilities of o1-like models QwQ, R1 and LIMO, and find that their performance can not be continuously improved through increasing CoT length.
    \item We reveal that insufficient self-revision capability of o1-like models is the primary reason for their failure in sequential scaling.
    \item We find that parallel scaling achieves better coverage and scalability than sequential revision for o1-like models. 
    \item Based on our insights into sequential and parallel scaling, we propose Shortest Majority Vote, a test-time scaling method that enhances majority voting by considering solution length, significantly outperforming traditional methods.
\end{enumerate}
\section{Related Work}
The success of o1 has ushered in a new scaling paradigm, \textbf{test-time compute scaling}, which enables continuous improvements in model performance by increasing computational expenditure during inference \citep{o1_blog,o1_system_card}. Currently, scaling test-time compute can be approached in two dimensions: parallel scaling and sequential scaling \citep{Scaling_test_time_compute,o1-roadmap}. 
\paragraph{Parallel Scaling}
Parallel scaling typicallly samples multiple solutions in parallel and pick one according to some guidence signal like reward. Notable examples of parallel scaling include Best-of-N Search \citep{OpenAIMathVerifierORM,Speculative_bon,BoNBoN,Variational_bon,BOND}, which is based on a reward model \citep{OpenAIMathVerifierORM,VerifySbyS}, and Majority Vote \citep{SC}, which exploits model uncertainty. The primary distinction between these approaches lies in the method used to select the final solution or answer after sampling multiple candidates. Both Best-of-N Search and Majority Vote are parallel scaling techniques at the solution level, while Tree-Search algorithms can be viewed as parallel scaling at the token or step level. Beam-Search \citep{TreeBoN,OVM,Self_evaluation_guided_beam_search,stochastic_beam_search} and MCTS \citep{RAP,Alpha-zero-like,AlphaMath,MCTS_for_code_generation} are classic examples of Tree-Search algorithms. All parallel scaling methods rely on guidance signals to select the optimal token, step, or solution from a set of candidates. 


\paragraph{Sequential Scaling}
Sequential scaling enhances test-time computation by generating progressively longer solutions along the sequence dimension. The most prevalent method of sequential scaling is Self-Revision, where \citet{Self_refine} first generate an initial response and then iteratively evaluate and refine it based on self-assessment. In contrast, \citet{self_debug,Critic} leverage external feedback—such as signals from a code execution environment—rather than self-evaluation to enhance solutions.

The effectiveness of sequential scaling with self-revision remains a contentious issue. \citet{LLM_cannot_self_correct,can-llm-self-revision} argue that models cannot achieve effective self-refinement without external feedback. Conversely, some researchers posit that evaluating a solution’s correctness is inherently easier than generating a correct solution \citep{janleike2022why_excited_about_AI_assisted_human_feedback}, suggesting that LLMs have the capacity for self-evaluation. \citet{SCoRE,teach-self-revision} show that it is possible to teach LLM to self-refine through reinforcement learning or supervised fine-tuning. \citet{Tree_search_vs_revisions} compared various test-time scaling algorithms and found that when feedback accuracy exceeds 90\%, Self-Revision outperforms Best-of-N Search.

\paragraph{o1-like Models} The release of o1 \citep{o1_blog,o1_system_card} has further underscored the significance of sequential scaling, as o1’s CoT length is substantially greater than that of conventional models. The research community has made significant efforts to reproduce the capabilities of o1 \citep{o1-journey1,o1-journey2,renda-1,renda-2,s1}, with QwQ \citep{qwq} and R1 \citep{deepseek-r1} and LIMO \citep{ye2025limoreasoning} emerging as the most successful attempts. However, Our findings reveal that for R1 and QwQ, extending solution length does not necessarily yield better performance due to the models’ limited self-revision capabilities. Parallel findings by \citet{o1-underthink} attribute this phenomenon to model underthinking, where models initially reach correct intermediate solutions but subsequently deviate toward incorrect conclusions during extended reasoning. \citet{o1-overthink,o1-pruner,o1-efficient} find that reducing the CoT length does not hurt the performance of o1-like models, which also supports our findings.

\section{Experiment Setting}
\paragraph{Models} Our experiments involved models from the QwQ \citep{qwq}, LIMO\citep{ye2025limoreasoning} and Deepseek-R1 series \citep{deepseek-r1}, including Deepseek-R1, Deepseek-R1-Distill-Qwen-32b, Deepseek-R1-Distill-Qwen-14b, and Deepseek-R1-Distill-Qwen-1.5b. For simplicy, we call these R1 models as R1-671b, R1-Distill-32b, R1-Distill-14b and R1-Distill-1.5b respectively. The models were run using SGLang framework \citep{sglang}, with the sampling temperature set to 0.7 and the maximum generation length set to 32k. We show the system prompt and instructions used for evaluation in Appendix \ref{app:prompt}.

\paragraph{Benchmark} We conducted comprehensive evaluations across four benchmarks: MATH-500 \citep{VerifySbyS}, AIME \citep{aime}, Omini-MATH \citep{omini-math}, and GPQA \citep{gpqa}. While MATH-500, AIME, and Omini-MATH focus on mathematical reasoning, GPQA encompasses broader scientific domains. For AIME evaluation, we utilized the AIMO validation set, comprising 90 questions from AIME 22, 23, and 24 \citep{aime}. Given the computational demands of evaluating the full Omini-MATH dataset (4.4K questions), we randomly sampled 500 questions to maintain efficiency. For GPQA, we focused on the diamond subset containing 198 questions. To ensure robust evaluation of answer correctness, we employed both the OpenCompass \citep{2023opencompass} and Qwen Math \citep{Qwen2.5-Math} evaluators, considering an answer correct if validated by either evaluator.

\begin{figure*}
    \centering
    \begin{subfigure}[b]{\textwidth}
        \includegraphics[width=\textwidth]{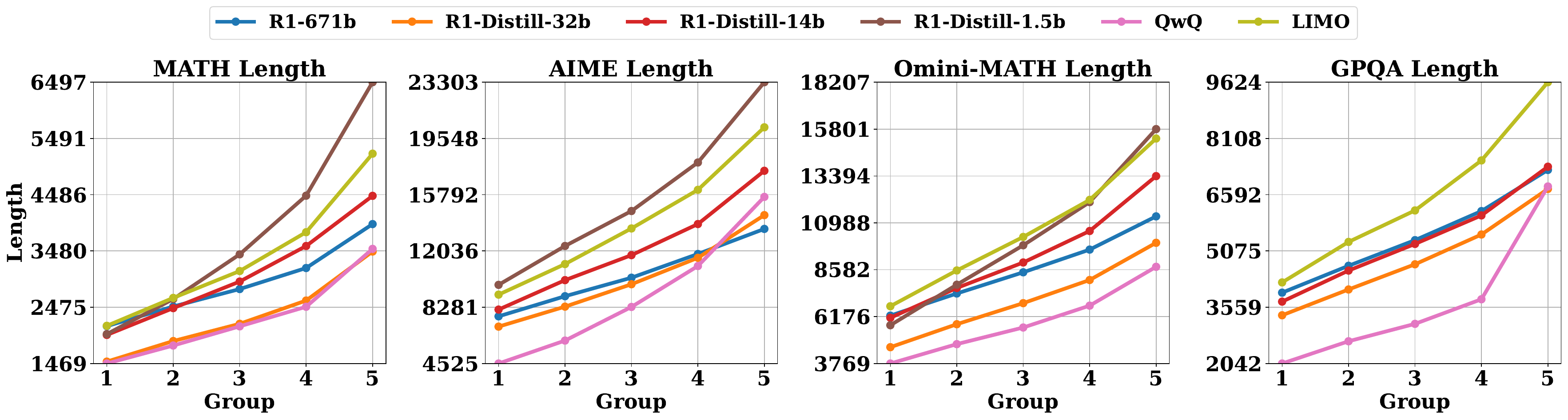}
        \caption{Evaluation for Solution length.}
        \label{fig:overall-len}
    \end{subfigure}
    
    \begin{subfigure}[b]{\textwidth}
        \includegraphics[width=\textwidth]{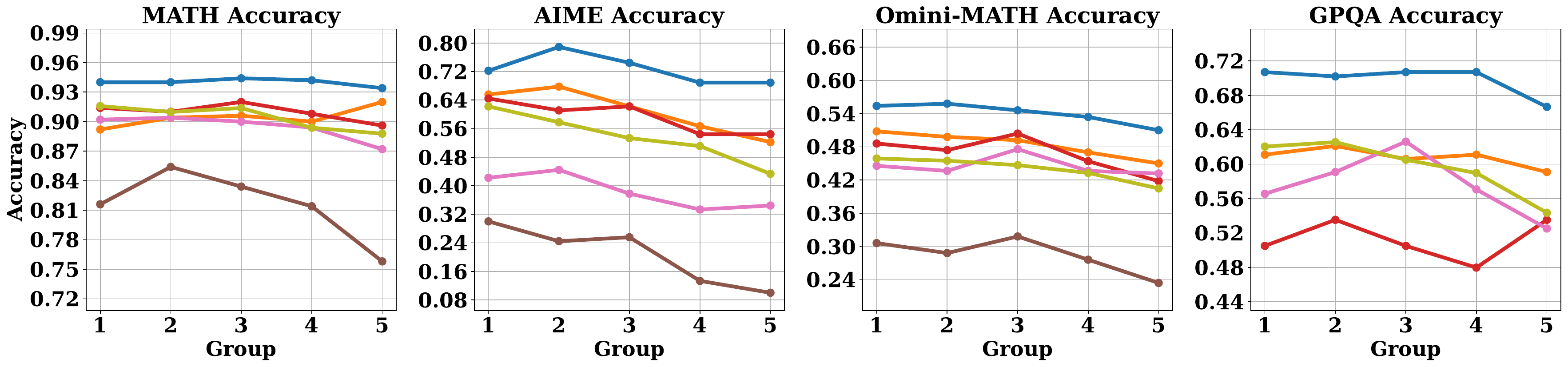}
        \caption{Evaluation for accuracy.}
        \label{fig:overall-acc}
    \end{subfigure}
    \caption{Solutions of QwQ and R1 were categorized into different groups according to their length and evaluated in terms of solution length (a) and accuracy (b). The categorization of solutions is progressed for each question independently, i.e., all groups of solutions are corresponding to the same questions.}
    
\end{figure*}
\section{The Failure of Sequential Scaling}

\subsection{Invalid Scaling of CoT Length: Longer CoTs Do not Improve Performance} \label{sec:invalid-scaling}
To investigate whether the accuracy of QwQ, R1 and LIMO genuinely improves with increasing CoT length, we sampled each model five times on the same question and sorted the five solutions by length in ascending order. We grouped the solutions based on their rank in this sorted list, with the $i$-th ranked solutions forming a distinct group. For instance, all the longest solutions (rank 5) from different questions formed one group, while all the shortest solutions (rank 1) formed another, resulting in 5 comprehensive solution groups for analysis.

We present the average lengths of the five groups of solutions in Figure \ref{fig:overall-len}. Since the grouping of solutions is based on their lengths, the differences in length between the groups are pronounced. The average length of the longest solutions is approximately twice that of the shortest solutions. This indicates that long-chain-of-thought (CoT) models like QwQ,  R1 and LIMO exhibit a high diversity in the lengths of the solutions they sample. 

There is no clear correlation between the length of solutions and the model's size. For example, R1-Distill-1.5b produces the longest solutions while QwQ (32b) generates the shortest. A comparison of solution lengths across different datasets shows that solutions for simpler datasets, such as Math, are significantly shorter than those for more difficult datasets, like AIME. This suggests that the model adjusts the solution length based on the difficulty of the problem.

The accuracy of the five groups of solutions is presented in Figure \ref{fig:overall-acc}. Although there is a significant disparity in solution lengths across the groups, the differences in accuracy are much less pronounced. Notably, we do not observe a consistent improvement in accuracy for either QwQ or R1 as solution length increases. This trend holds true across all model variants as well as across all evaluated datasets. In some cases, we even observe an inverse scaling phenomenon, where accuracy decreases with increasing CoT length, especially on more difficult datasets like AIME and Omini-MATH. These findings cast doubt on the presumed test-time scaling capabilities of o1-like models, challenging the assumption that extended reasoning chains inherently yield superior problem-solving performance. 

To make the relationship between  CoT length and accuracy more clear, we compared the lengths of correct and incorrect solutions for the same question. First, we identified questions that had both correct and incorrect answers. For each of these questions, we calculated the average length of correct and incorrect solutions. We then averaged these values across all questions to determine the overall average length for correct and incorrect solutions. The results are shown in Figure \ref{fig:overall-correct-vs-incorrect}. We found that, for QwQ, R1 and LIMO, across all model sizes and datasets, the length of correct solutions is consistently shorter than that of incorrect solutions. This observation suggests that longer CoTs do not necessarily lead to better performance and may even be associated with lower accuracy. Moreover, we observed that for weaker models, such as QwQ and R1-Distill-1.5B, the gap in solution length between correct and incorrect solutions is significantly larger than for stronger models, such as R1-671b. This suggests that the invalid scaling phenomenon is more pronounced in the weaker models.

\subsection{Explaining Invalid Scaling: The Key Factor is the Failure of Self-Revision}
\begin{figure*}[t]
    \centering
    \begin{subfigure}[b]{0.4\textwidth}
        \includegraphics[width=\textwidth]{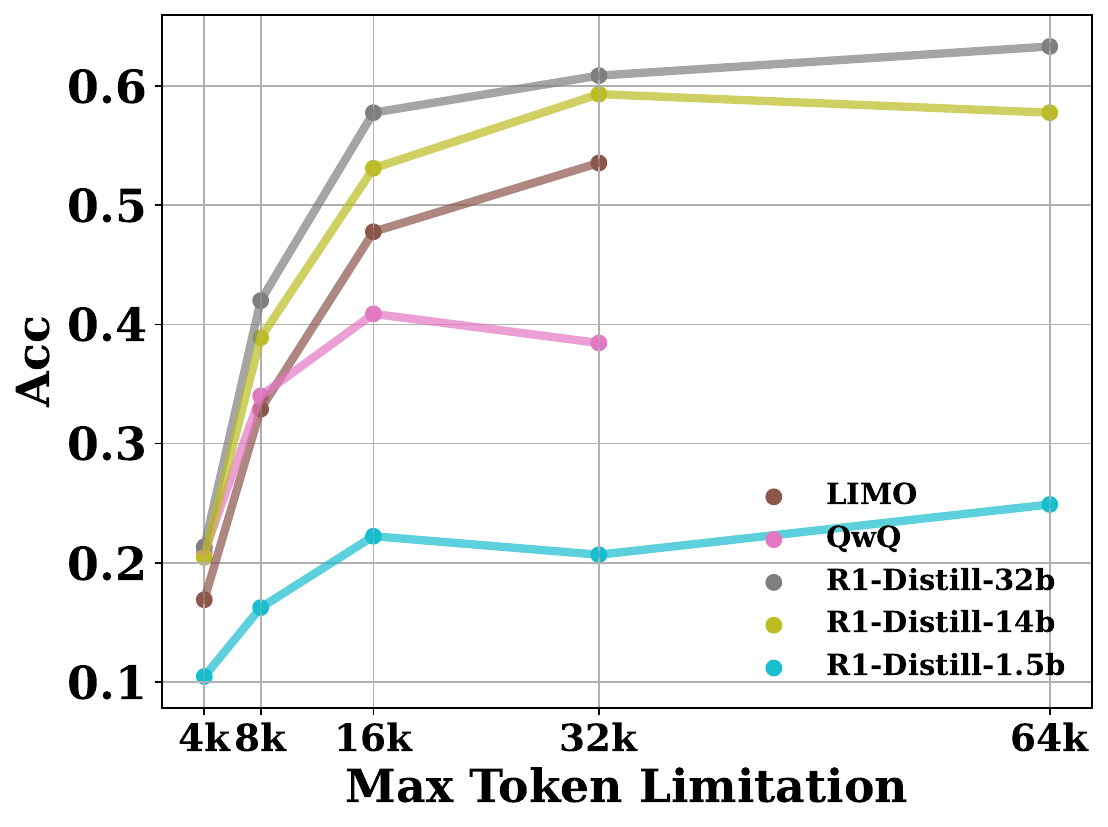}
        \caption{Max Token Limitation}
        \label{fig:max-token-limit}
    \end{subfigure}
    \begin{subfigure}[b]{0.4\textwidth}
        \includegraphics[width=\textwidth]{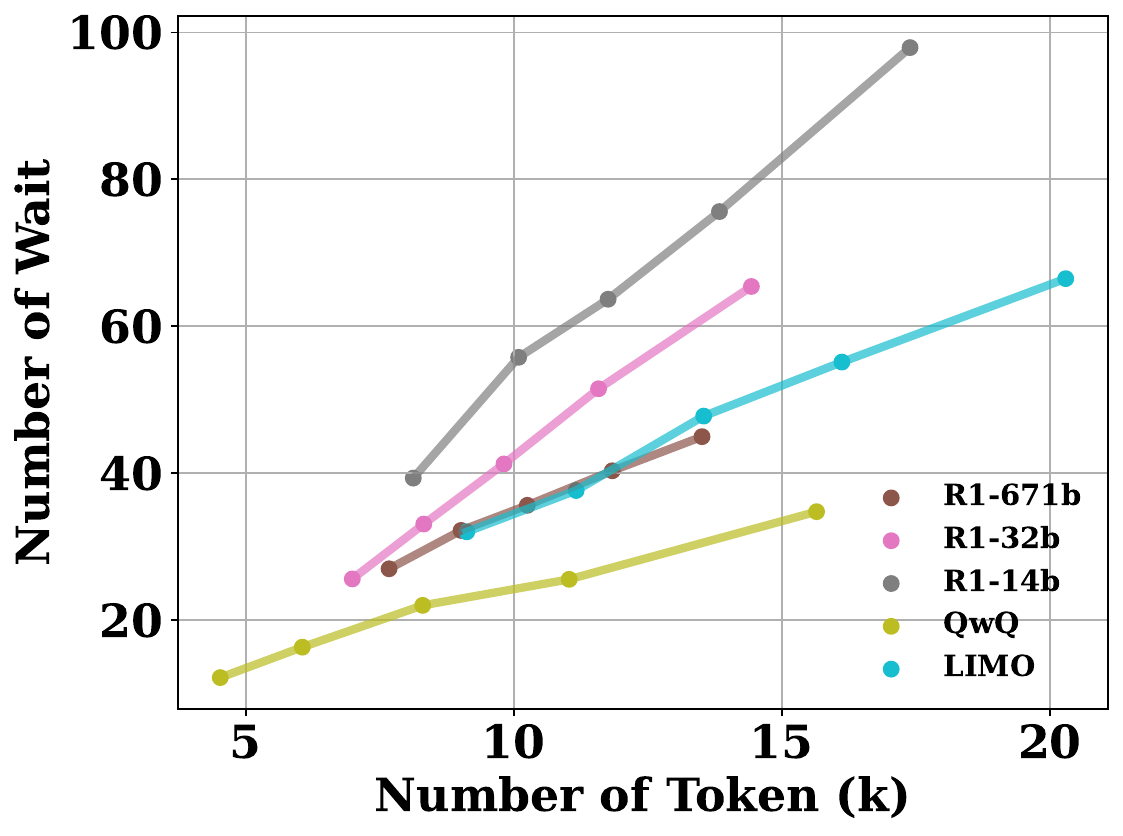}
        \caption{Frequence of ``Wait''}
        \label{fig:wait-count}
    \end{subfigure}
    \caption{(a): The relationship between model accuracy and the generation parameter Max Token Limitation. (b): The relationship between solution length and the average number of ``wait'' occur in a solution.}
    \label{fig:enter-label}
\end{figure*}


In Section \ref{sec:invalid-scaling}, we observed the phenomenon that long solutions exhibit lower accuracy compared to short solutions. In this section, we investigate the underlying reasons for this phenomenon. We first analyzed how the maximum token limitation affects generation performance and confirmed that the observed invalid scaling phenomenon was not caused by constraints in the maximum token length. Next, we examined the differences between long and short solutions, finding that long solutions exhibit a higher frequency of self-revision. Moreover, our analysis suggests a strong correlation between self-revision, solution length, and accuracy.

\paragraph{Max Token Limitation}

The max token limitation parameter controls the maximum number of tokens a model can generate for a question, which plays a critical role in influencing model accuracy, especially when generating long solutions. To explore its impact, we tested several max token limitation values and compared the performance of QwQ, R1 and LIMO on the AIME benchmark. The results are shown in Figure \ref{fig:max-token-limit}, which revealed that 16k is a key threshold: when the max token limitation is below this value, it significantly affects the model performance. However, increasing the max token limitation beyond 16k leads to diminishing returns, particularly for QwQ. In our other experiments, we set the max token limitation to 32k, suggesting that this parameter is not the main cause of invalid scaling.

\paragraph{Difference between Short and Long CoT}
To understand why long solutions of QwQ, R1 and LIMO is not better than short solutions, we analyzed their differences. We observed that QwQ, R1 and LIMO all primarily extend solution length through self-revision, characterized by markers such as ``Wait'' and ``Alternatively''. We show some examples of that in Appendix \ref{app:revision-examples}. To quantify this phenomenon, we counted the occurrences of ``wait'' in solutions of QwQ, R1 and LIMO in Figure \ref{fig:wait-count}. The results demonstrates a strong linear correlation between solution length and the frequency of self-correction markers for all models. This suggests that the mechanisms of self-revision may play a significant role in generating longer solutions. 
\begin{figure*}[t]
    \centering
    \begin{subfigure}[b]{0.32\textwidth}
        \includegraphics[width=\textwidth]{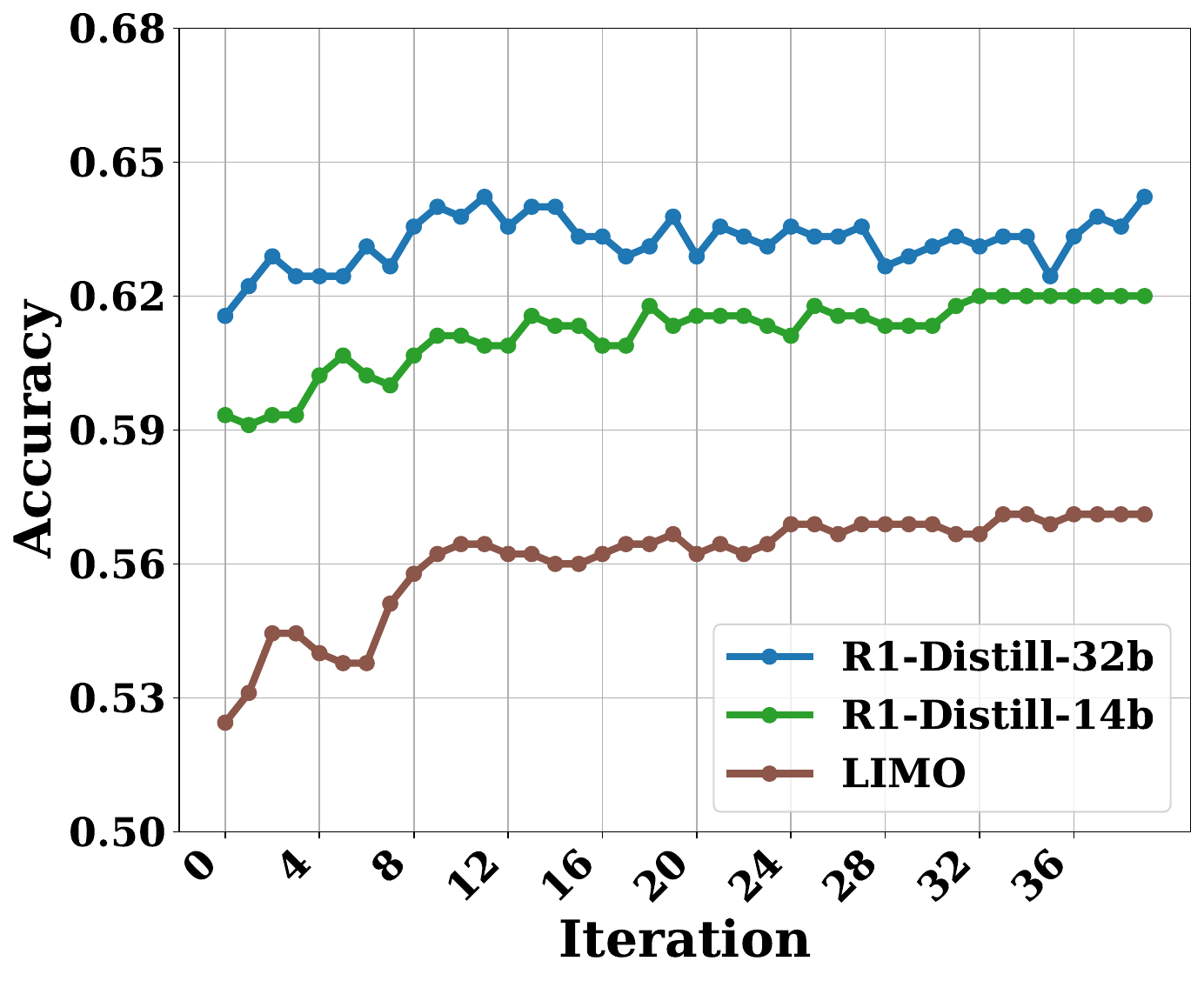}
        \caption{Acc of R1-Distill-32b, -14b}
        \label{fig:32b-14b-acc}
    \end{subfigure}
    \hfill
    \begin{subfigure}[b]{0.32\textwidth}
        \includegraphics[width=\textwidth]{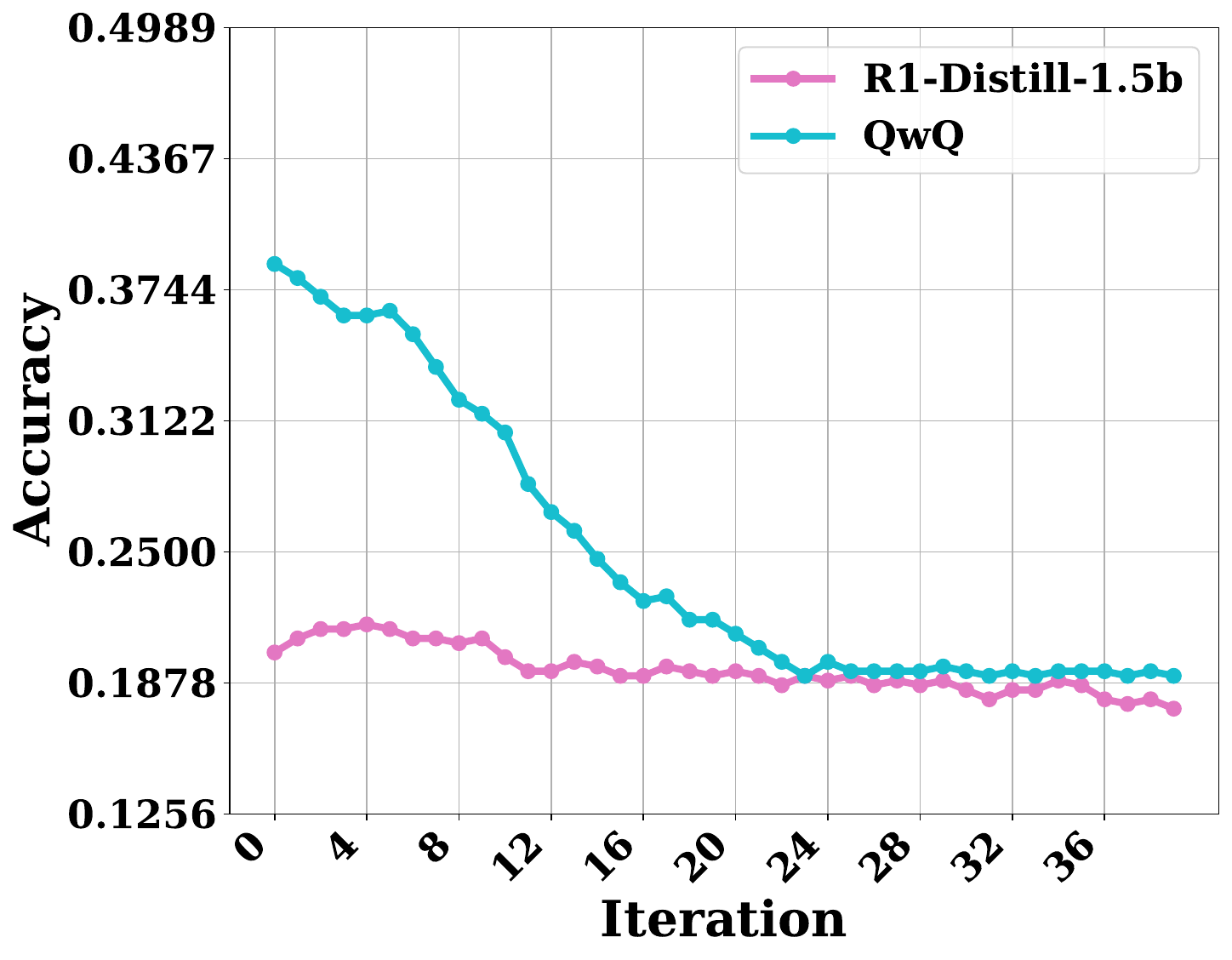}
        \caption{Acc of R1-Distill-1.5b, QwQ}
        \label{fig:1.5b-qwq-acc}
    \end{subfigure}
    \hfill
    \begin{subfigure}[b]{0.32\textwidth}
        \includegraphics[width=\textwidth]{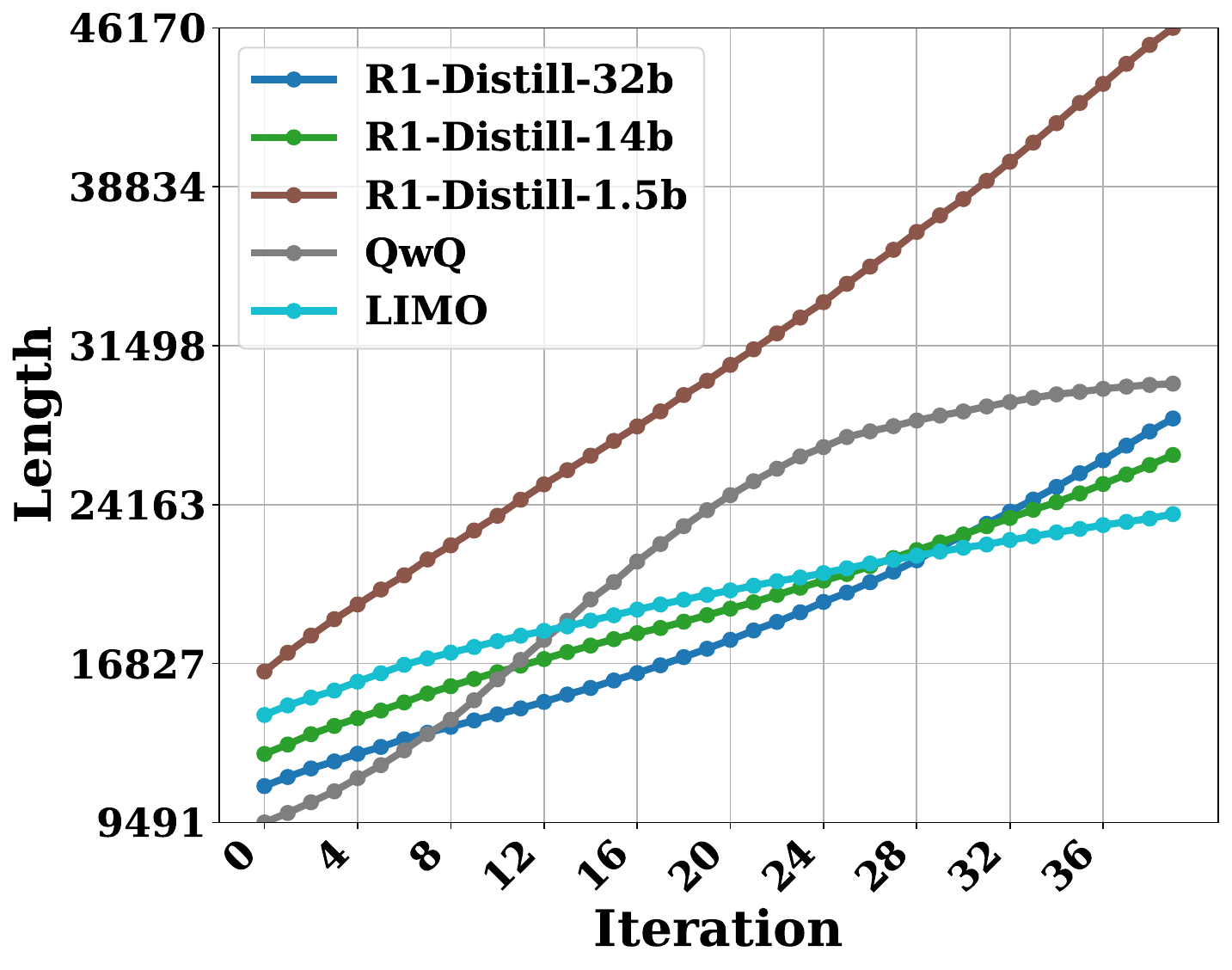}
        \caption{Solution lengths.}
        \label{fig:increased-len}
    \end{subfigure}
    \caption{(a): Accuracy of R1-Distill-32b, R1-Distill-14b and LIMO during sequential revisions. (b): Accuracy of R1-Distill-1.5b and QwQ during sequential revisions. (c) Solution length increased with the more revision steps.}
\end{figure*}

\begin{figure*}[t]
    \centering
    \includegraphics[width=1.0\linewidth]{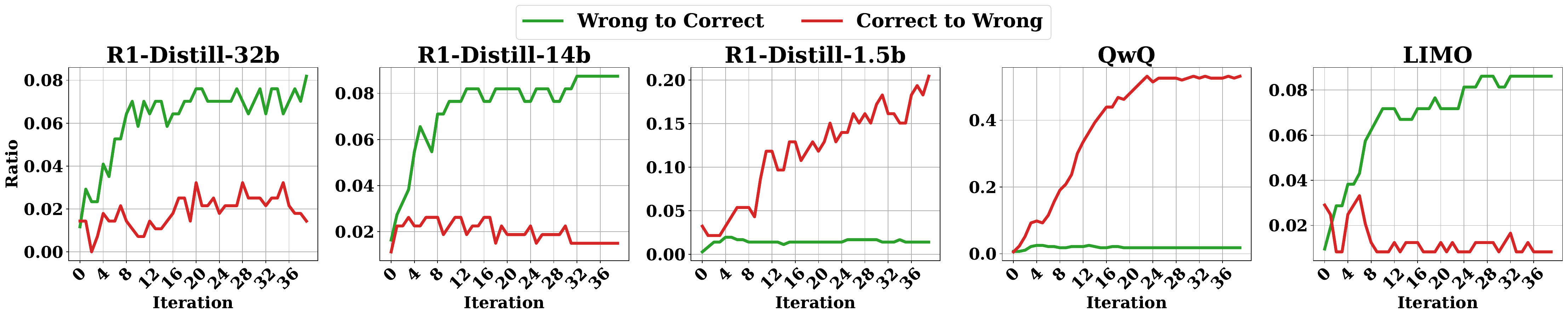}
    \caption{The ratio of turning an initial correct answer to incorrect one (correct to wrong) and an initial incorrect answer to a correct one (wrong to correct)  during sequential scaling.}
    \label{fig:revision}
\end{figure*}
\paragraph{Scaling Solution Length with Self-Revision}\label{sec:sequential-revision}
We have tried to investigate the revision behaviors inside the sampled solutions, however, it is difficult to extract the initial solution and the following revision exactly from QwQ, R1 and LIMO's solutions. Alternatively to that, we prompted the models to continue thinking based on their sampled solutions. 

QwQ, R1 and LIMO often conclude their solutions with phrases like ``final answer: ...'', and R1 additionally outputs a `</think>' tag followed by a final response. To facilitate smoother continuation of the reasoning process, we removed the ``final answer'' portion from the solutions. We then used the keyword ``Wait'' or ``Alternatively'' as the prompt to encourage self-revision. We calculated the probabilities of the model predicting the next token as ``Wait'' or ``Alternatively'' and selected the one with the higher probability as the prompt.

We prompted QwQ, R1 and LIMO to continue reasoning for 40 additional steps on the AIME benchmark. We show the results in Figure \ref{fig:increased-len}, from which we observe that the solution length increase almost linearly with additional steps. After 40 steps, the solution length of QwQ and R1 is almost third as their original length.

We show the accuracy after sequential revision in Figure \ref{fig:32b-14b-acc} and \ref{fig:1.5b-qwq-acc}. Our results reveal that the accuracy of QwQ and R1-Distill-1.5b decreases constantly as the number of reasoning steps increases, while the accuracy of R1-Distill-32b, R1-Distill-14b and LIMO initially improves and then oscillates with further reasoning steps. Further analysis in Appendix \ref{app:short-long-revision} reveal that the improvement on R1-Distill-32b, R1-Distill-14b and LIMO during revisions mainly comes from the revision on short solutions. These results corroborate our previous experimental findings, suggesting that longer solutions do not improve performance, especially for weaker models such as QwQ and R1-Distill-1.5b. These findings suggest that the reason why longer solutions do not consistently lead to better performance in QwQ, R1 and LIMO may lie in the failure of self-revision.

\paragraph{Investigating Self-Revision Behavior}
To further investigate the effectiveness of self-revision, we analyzed the proportion of cases where the model corrected an initial incorrect answer to a correct one versus changing an initial correct answer to an incorrect one during scaling solution length. We found that, the proportions of changing a incorrect answer to an correct one is extremely low, always below 10\%. Notably, for QwQ and R1-Distill-1.5b, the proportion of changing a correct answer to an incorrect one was even higher than that of correcting an incorrect answer to a correct one. This observation helps explain why prompting QwQ and R1-Distill-1.5b to continue reasoning led to a decrease in accuracy. For simplicty, we call the proportions of changing a incorrect answer to an correct one as the successful-revision rate, while the reverse as the failed-revision rate.

Although R1-Distill-32b, R1-Distill-14b and LIMO exhibit a higher successful-revision rate than failed-revision rate, the increase of successful-revision rate plateaus after approximately 10 steps, with further revisions providing no additional benefits. This observation explains why their accuracy during sequential scaling initially increases with multiple rounds of revision but later stabilizes with fluctuations.

\begin{table}[t]
    \centering
    \begin{tabular}{ccccl}
        \toprule
        R1-32b & R1-14b & R1-1.5b & QwQ  &LIMO\\
        \midrule
         72\% & 70\% & 58\% & 32\%  & 54\%\\
        \bottomrule
    \end{tabular}
    \caption{The proportion of the revisions that models sitck to the original wrong answers.}
    \label{tab:stick}
\end{table}

\begin{figure*}[t]
    \centering
    \begin{subfigure}[b]{0.4\textwidth}
        \includegraphics[width=\textwidth]{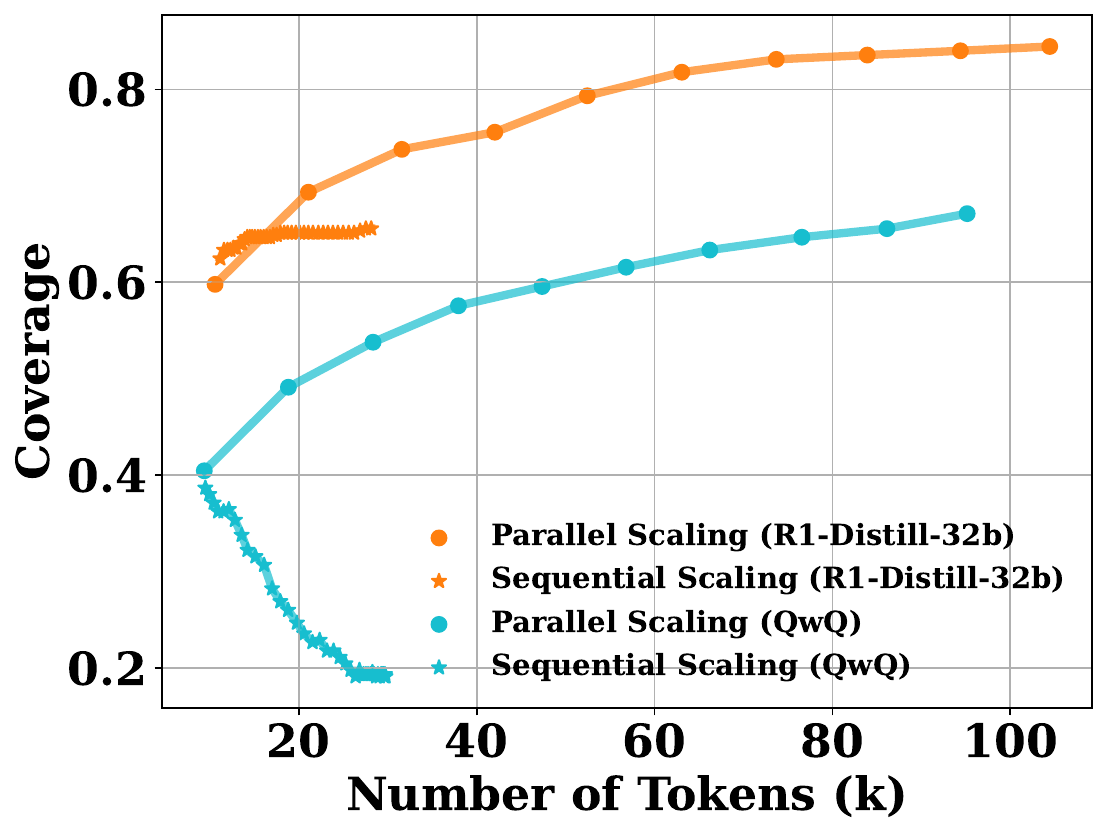}
        \caption{Evaluation on Coverage.}
        \label{fig:coverage-seq-vs-parallel}
    \end{subfigure}
    \begin{subfigure}[b]{0.4\textwidth}
        \includegraphics[width=\textwidth]{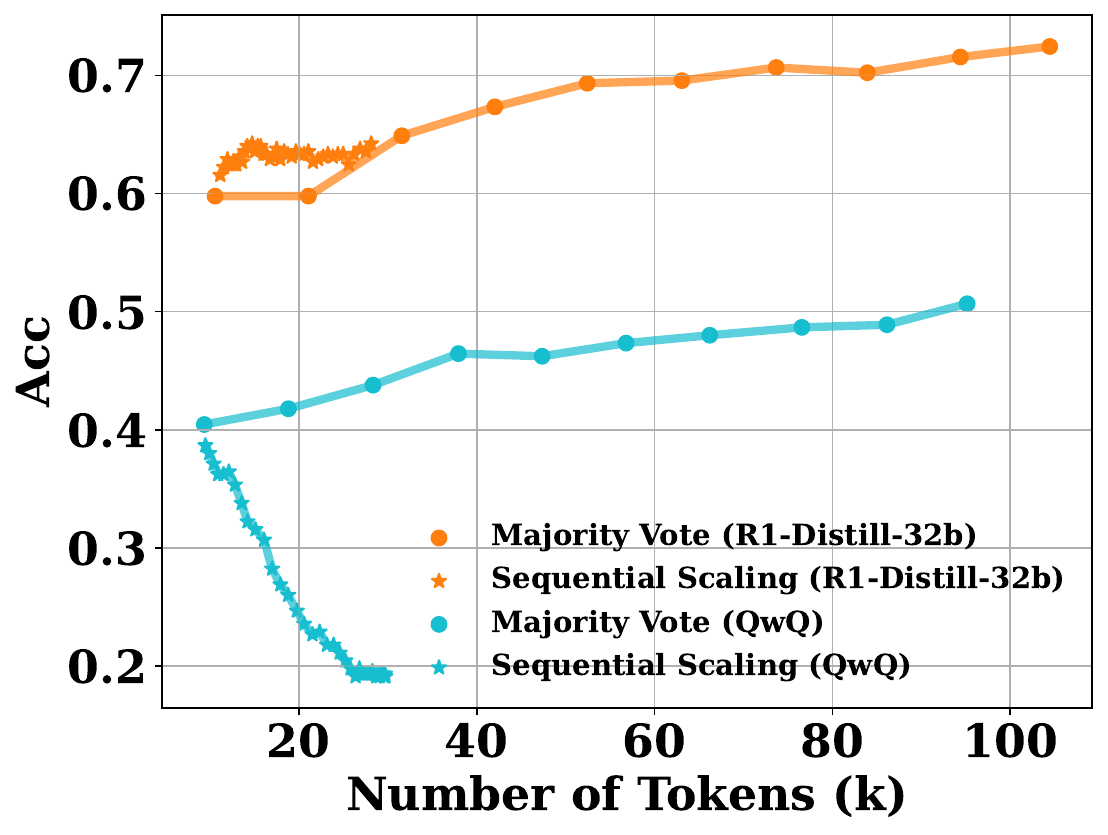}
        \caption{Evaluation on Accuracy}
        \label{fig:acc-seq-vs-parallel}
    \end{subfigure}
    \caption{(a): the coverage of sequential scaling and parallel scaling on AIME. (b): the accuracy of squential revision and majority vote on AIME.}
\end{figure*}
The successful-revision rate of QwQ, R1 and LIMO are all below 10\%, what is the outcome of the model’s self-revision in unsuccessful cases? We hypothesize that, in most instances, the model simply keeps its original answer unchanged. To validate that, we computed the proportion of instances where the model persists with its original answer, even when it is incorrect, and the results were as expected. As shown in Figure \ref{fig:revision}, when the original answer is wrong, both R1-Distill-32b and R1-Distill-14b maintain the original answer in over 70\% of cases. Although retaining the original answer does not reduce accuracy, it also makes the scaling solution length ineffective. This phenomenon suggests that the model’s ability to early stop may also be a critical factor influencing whether its performance improves with an increasing solution length.

The above analysis indicates that the key factor determining whether o1-like models' performance improve with an increase in solution length is their ability to self-revise. The model's accuracy increases with the more incorrect answers revised to correct and vice versa.

\begin{figure*}[t]
    \centering
    \includegraphics[width=\textwidth]{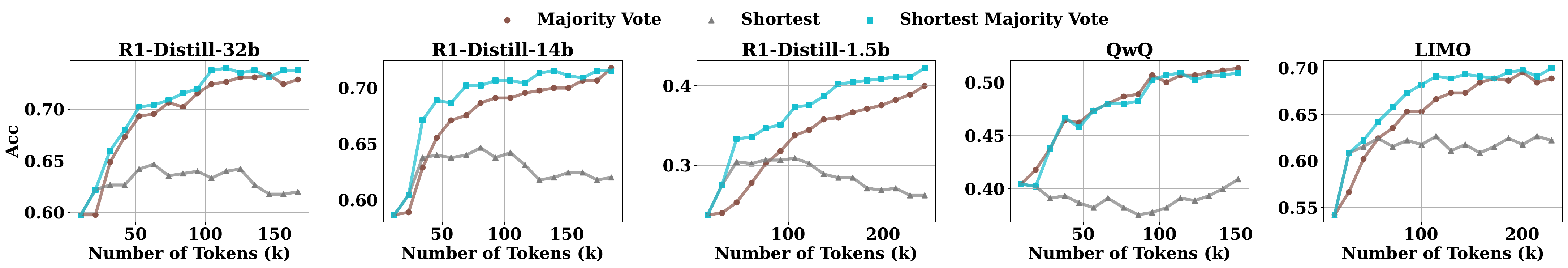}
    \caption{Parallel-scaling performance of Majority Vote, Shortest and Shortest Majority Vote on AIME.}
    \label{fig:short-majority-vote}
\end{figure*}

\section{Sequential Scaling vs. Parallel Scaling}\label{sec:seq-vs-parallel}
Based on our experimental findings presented in Section \ref{sec:sequential-revision}, sequential scaling demonstrates limited effectiveness for QwQ, R1 and LIMO. An alternative approach to scaling test-time compute is parallel scaling, which generates multiple solutions in parallel and selects the best one as the final answer.  

We compared the performance of sequential scaling and parallel scaling in terms of the coverage (pass@k score) and accuracy of QwQ and R1, which are shown in Figure \ref{fig:coverage-seq-vs-parallel} and \ref{fig:acc-seq-vs-parallel} respectively. For sequential scaling, we iteratively prompt models to self-revise for 40 steps. While for parallel scaling, we parallely sample 10 solutions. The coverage is evaluated by counting the proportion of whether multiple candidate answers contain a correct one. In parallel scaling, coverage increases by one if at least one sampled solution is correct. Similarly, in sequential scaling, coverage increases by one if at least one revision iteration succeeds.

Our findings show that, for the same number of generated tokens, parallel scaling provides a significantly larger improvement in coverage compared to sequential scaling, for both R1-Distill-32b and QwQ. However, a practical parallel scaling method must select a final answer from a set of candidate answers. We implement parallel scaling using majority vote \citep{SC} and sequential scaling by taking the answer from the last revision as the final answer. Since majority voting requires at least three solutions to be effective, it does not provide any benefit when scaling the number of solutions from 1 to 2. In contrast, sequential revision is effective for R1-Distill-32b when scaling the number of tokens to 10k, but further scaling does not yield additional benefits. Additionally, because sequential scaling involves attention over a longer context, its computational cost is much higher than that of parallel scaling when generating the same number of tokens. 


\section{Application of Our Findings: Shortest Majority Vote}\label{sec:new-method}
\begin{table*}[t]
    \centering
    \resizebox{0.9\textwidth}{!}{
    \begin{tabular}{cccccccc}
    \toprule
 \multirow{2}{*}{Model} & \multirow{2}{*}{Solutions} & \multicolumn{3}{c}{AIME} & \multicolumn{3}{c}{GPQA}\\
        \cmidrule{3-8}
        & & MV& Shortest& Shortest MV& MV& Shortest&Shortest MV\\
        \midrule
R1-Distill-32b&\multirow{5}{*}{2}& 59.77&\textbf{62.22}& \textbf{62.22}& 61.41& \textbf{62.52}&\textbf{62.52}\\
R1-Distill-14b & & 58.88&\textbf{60.44}& \textbf{60.44}& 51.21& \textbf{52.32}&\textbf{52.32}\\
R1-Distill-1.5b & & 24&\textbf{27.55}& \textbf{27.55} & 15.25& \textbf{15.35}&\textbf{15.35}\\
QwQ & & \textbf{41.77}&40.22& 40.22& \textbf{58.05}& 57.02&57.02\\
 LIMO& & 56.66& \textbf{60.88}& \textbf{60.88}& 50.46& \textbf{54.56}&\textbf{54.56}\\
\midrule
R1-Distill-32b & \multirow{5}{*}{16}& 72.88&61.99& \textbf{73.77}& 63.33& 61.21&\textbf{63.53}\\
R1-Distill-14b & & \textbf{71.77}&62.00& 71.55& 56.16& \textbf{56.66}&56.46\\
R1-Distill-1.5b & & 40.00&26.22& \textbf{42.22}& 29.59& 27.77&\textbf{30.20}\\
QwQ  & & \textbf{51.33}&40.88& 50.88& \textbf{62.25}& 56.82&\textbf{62.25}\\
 LIMO& & 68.88& 62.22& \textbf{70.00}& 55.58& 50.15&\textbf{55.89}\\
  \bottomrule
    \end{tabular}}
    \caption{Performance comparison between Majority Vote (MV), Shortest and Shortest Majority Vote (Shortest MV) on AIME and GPQA, when there are 2 and 16 solutions sampled.}
    \label{tab:shortest-mv}
\end{table*}
\label{Application of Our Findings: Shortest Majority Vote}
Given the limitation of sequential scaling of the current o1-like models, we turn to parallel scaling techniques and incorperate it with our insight on sequential scaling. Specifically, we propose a new Parallel Scaling algorithm: Shortest Majority Vote. Shortest Majority Vote is an extension of Majority Vote, but it accounts for the length of the solutions generated by the model. In the original Majority Vote, solutions with the same answer are grouped into a single category, and the number of solutions in each category is counted, with the answer corresponding to the category with the most solutions selected as the final answer. In contrast, Shortest Majority Vote not only counts the number of solutions in each category, but also computes the average length of the solutions in each category. Let the number of solutions in the $i$-th category be $c_i$ and the average solution length in that category be $l_i$. The score for category $i$ in Shortest Majority Vote is computed as:
\begin{equation}
    s_i=\frac{c_i}{\log{l_i}}
\end{equation}
and the final answer is chosen from the category with the highest score. The score $s_i$ is designed with the assumption that the correct answer is more likely to appear in categories with a larger number of solutions and shorter solution lengths. Shortest Majority Vote offers two key advantages: first, it is particularly effective for some o1-like models, where performance deteriorates with increasing solution length; second, it enables the use of solution length as a guidence signal for identifying superior solutions when candidate solutions are limited, especially in cases where conventional Majority Vote becomes ineffective due to having only two candidate solutions.


We evaluated the performance of Shortest Majority Vote and Majority Vote through experiments on the AIME and GPQA benchmarks, sampling 16 solutions from QwQ, R1 and LIMO models. We implemented a simple baseline approach, denoted as "Shortest," which selects the answer from the solution with the minimal length. The experimental results are presented in Table \ref{tab:shortest-mv} and Figure \ref{fig:short-majority-vote}. Table \ref{tab:shortest-mv} demonstrates that Shortest Majority Vote  significantly outperforms both Majority Vote and Shortest methods, particularly on the AIME benchmark. Figure \ref{fig:short-majority-vote} illustrates the parallel-scaling performance of these three methods, showing that as the number of generated tokens increases, Shortest Majority Vote maintains superior performance over both alternatives on AIME. The corresponding parallel-scaling results for GPQA are provided in Appendix \ref{app:gpqa-scale}. Notably, while Shortest performs better than Majority Vote when only two solutions are sampled, it exhibits inferior performance in all other scenarios. These empirical findings strongly support the effectiveness of the Shortest Majority Vote approach.

\section{Conclusion}
In this study, we challenged the assumption that o1-like models like QwQ and R1 models have test-time scaling capability. 
We found that shorter solutions often outperform longer ones, and that sequential scaling through self-revision has limited effectiveness. Based on these insights, we developed Shortest Majority Vote, a parallel scaling method that considers solution length, which significantly outperformed traditional majority vote.

\section*{Limitations}
\begin{enumerate}
    \item Given the considerable cost of R1-671b, evaluation on it was limited to the experiments in Figures 1 and 2, whereas distilled R1 was utilized for all subsequent.
    \item Our experimental framework was limited to static model checkpoints. Future research should investigate test-time scaling behavior using dynamic checkpoints in reinforcement learning settings.

    \item While the proposed shortest majority method may have limited applicability for models with strong sequential-scaling capabilities, solution length remains a valuable guidance signal for candidate selection in parallel scaling scenarios. The method can be adapted to a Longest Majority Vote variant for such cases.
\end{enumerate}

\bibliography{colm2024_conference}

\begin{thebibliography}{47}
\providecommand{\natexlab}[1]{#1}
\providecommand{\url}[1]{\texttt{#1}}
\expandafter\ifx\csname urlstyle\endcsname\relax
  \providecommand{\doi}[1]{doi: #1}\else
  \providecommand{\doi}{doi: \begingroup \urlstyle{rm}\Url}\fi

\bibitem[AIMO(2018)]{aime}
AIMO.
\newblock Dataset card for aimo validation aime.
\newblock \url{https://huggingface.co/datasets/AI-MO/aimo-validation-aime}, 2018.

\bibitem[Amini et~al.(2024)Amini, Vieira, and Cotterell]{Variational_bon}
Afra Amini, Tim Vieira, and Ryan Cotterell.
\newblock Variational best-of-n alignment.
\newblock \emph{CoRR}, abs/2407.06057, 2024.
\newblock \doi{10.48550/ARXIV.2407.06057}.
\newblock URL \url{https://doi.org/10.48550/arXiv.2407.06057}.

\bibitem[Arora \& Zanette(2025)Arora and Zanette]{o1-efficient}
Daman Arora and Andrea Zanette.
\newblock Training language models to reason efficiently.
\newblock 2025.
\newblock URL \url{https://api.semanticscholar.org/CorpusID:276235717}.

\bibitem[Chen et~al.(2024{\natexlab{a}})Chen, Liao, Li, and Fan]{AlphaMath}
Guoxin Chen, Minpeng Liao, Chengxi Li, and Kai Fan.
\newblock Alphamath almost zero: process supervision without process.
\newblock \emph{CoRR}, abs/2405.03553, 2024{\natexlab{a}}.
\newblock \doi{10.48550/ARXIV.2405.03553}.
\newblock URL \url{https://doi.org/10.48550/arXiv.2405.03553}.

\bibitem[Chen et~al.(2024{\natexlab{b}})Chen, Xu, Liang, He, Pang, Yu, Song, Liu, Zhou, Zhang, Wang, Tu, Mi, and Yu]{o1-overthink}
Xingyu Chen, Jiahao Xu, Tian Liang, Zhiwei He, Jianhui Pang, Dian Yu, Linfeng Song, Qiuzhi Liu, Mengfei Zhou, Zhuosheng Zhang, Rui Wang, Zhaopeng Tu, Haitao Mi, and Dong Yu.
\newblock Do not think that much for 2+3=? on the overthinking of o1-like llms.
\newblock \emph{ArXiv}, abs/2412.21187, 2024{\natexlab{b}}.
\newblock URL \url{https://api.semanticscholar.org/CorpusID:275133600}.

\bibitem[Chen et~al.(2024{\natexlab{c}})Chen, Lin, Sch{\"{a}}rli, and Zhou]{self_debug}
Xinyun Chen, Maxwell Lin, Nathanael Sch{\"{a}}rli, and Denny Zhou.
\newblock Teaching large language models to self-debug.
\newblock In \emph{The Twelfth International Conference on Learning Representations, {ICLR} 2024, Vienna, Austria, May 7-11, 2024}. OpenReview.net, 2024{\natexlab{c}}.
\newblock URL \url{https://openreview.net/forum?id=KuPixIqPiq}.

\bibitem[Chen et~al.(2024{\natexlab{d}})Chen, White, Mooney, Payani, Su, and Sun]{Tree_search_vs_revisions}
Ziru Chen, Michael White, Raymond~J. Mooney, Ali Payani, Yu~Su, and Huan Sun.
\newblock When is tree search useful for {LLM} planning? it depends on the discriminator.
\newblock In Lun{-}Wei Ku, Andre Martins, and Vivek Srikumar (eds.), \emph{Proceedings of the 62nd Annual Meeting of the Association for Computational Linguistics (Volume 1: Long Papers), {ACL} 2024, Bangkok, Thailand, August 11-16, 2024}, pp.\  13659--13678. Association for Computational Linguistics, 2024{\natexlab{d}}.
\newblock \doi{10.18653/V1/2024.ACL-LONG.738}.
\newblock URL \url{https://doi.org/10.18653/v1/2024.acl-long.738}.

\bibitem[Cobbe et~al.(2021)Cobbe, Kosaraju, Bavarian, Chen, Jun, Kaiser, Plappert, Tworek, Hilton, Nakano, Hesse, and Schulman]{OpenAIMathVerifierORM}
Karl Cobbe, Vineet Kosaraju, Mohammad Bavarian, Mark Chen, Heewoo Jun, Lukasz Kaiser, Matthias Plappert, Jerry Tworek, Jacob Hilton, Reiichiro Nakano, Christopher Hesse, and John Schulman.
\newblock Training verifiers to solve math word problems.
\newblock \emph{CoRR}, abs/2110.14168, 2021.
\newblock URL \url{https://arxiv.org/abs/2110.14168}.

\bibitem[Contributors(2023)]{2023opencompass}
OpenCompass Contributors.
\newblock Opencompass: A universal evaluation platform for foundation models.
\newblock \url{https://github.com/open-compass/opencompass}, 2023.

\bibitem[DeepSeek-AI et~al.(2025)DeepSeek-AI, Guo, Yang, Zhang, Song, Zhang, Xu, Zhu, Ma, Wang, Bi, Zhang, Yu, Wu, Wu, Gou, Shao, Li, Gao, Liu, Xue, Wang, Wu, Feng, Lu, Zhao, Deng, Zhang, Ruan, Dai, Chen, Ji, Li, Lin, Dai, Luo, Hao, Chen, Li, Zhang, Bao, Xu, Wang, Ding, Xin, Gao, Qu, Li, Guo, Li, Wang, Chen, Yuan, Qiu, Li, Cai, Ni, Liang, Chen, Dong, Hu, Gao, Guan, Huang, Yu, Wang, Zhang, Zhao, Wang, Zhang, Xu, Xia, Zhang, Zhang, Tang, Li, Wang, Li, Tian, Huang, Zhang, Wang, Chen, Du, Ge, Zhang, Pan, Wang, Chen, Jin, Chen, Lu, Zhou, Chen, Ye, Wang, Yu, Zhou, Pan, Li, Zhou, Wu, Ye, Yun, Pei, Sun, Wang, Zeng, Zhao, Liu, Liang, Gao, Yu, Zhang, Xiao, An, Liu, Wang, Chen, Nie, Cheng, Liu, Xie, Liu, Yang, Li, Su, Lin, Li, Jin, Shen, Chen, Sun, Wang, Song, Zhou, Wang, Shan, Li, Wang, Wei, Zhang, Xu, Li, Zhao, Sun, Wang, Yu, Zhang, Shi, Xiong, He, Piao, Wang, Tan, Ma, Liu, Guo, Ou, Wang, Gong, Zou, He, Xiong, Luo, You, Liu, Zhou, Zhu, Xu, Huang, Li, Zheng, Zhu, Ma, Tang, Zha, Yan, Ren, Ren, Sha, Fu, Xu, Xie, Zhang,
  Hao, Ma, Yan, Wu, Gu, Zhu, Liu, Li, Xie, Song, Pan, Huang, Xu, Zhang, and Zhang]{deepseek-r1}
DeepSeek-AI, Daya Guo, Dejian Yang, Haowei Zhang, Junxiao Song, Ruoyu Zhang, Runxin Xu, Qihao Zhu, Shirong Ma, Peiyi Wang, Xiao Bi, Xiaokang Zhang, Xingkai Yu, Yu~Wu, Z.~F. Wu, Zhibin Gou, Zhihong Shao, Zhuoshu Li, Ziyi Gao, Aixin Liu, Bing Xue, Bingxuan Wang, Bochao Wu, Bei Feng, Chengda Lu, Chenggang Zhao, Chengqi Deng, Chenyu Zhang, Chong Ruan, Damai Dai, Deli Chen, Dongjie Ji, Erhang Li, Fangyun Lin, Fucong Dai, Fuli Luo, Guangbo Hao, Guanting Chen, Guowei Li, H.~Zhang, Han Bao, Hanwei Xu, Haocheng Wang, Honghui Ding, Huajian Xin, Huazuo Gao, Hui Qu, Hui Li, Jianzhong Guo, Jiashi Li, Jiawei Wang, Jingchang Chen, Jingyang Yuan, Junjie Qiu, Junlong Li, J.~L. Cai, Jiaqi Ni, Jian Liang, Jin Chen, Kai Dong, Kai Hu, Kaige Gao, Kang Guan, Kexin Huang, Kuai Yu, Lean Wang, Lecong Zhang, Liang Zhao, Litong Wang, Liyue Zhang, Lei Xu, Leyi Xia, Mingchuan Zhang, Minghua Zhang, Minghui Tang, Meng Li, Miaojun Wang, Mingming Li, Ning Tian, Panpan Huang, Peng Zhang, Qiancheng Wang, Qinyu Chen, Qiushi Du, Ruiqi Ge, Ruisong
  Zhang, Ruizhe Pan, Runji Wang, R.~J. Chen, R.~L. Jin, Ruyi Chen, Shanghao Lu, Shangyan Zhou, Shanhuang Chen, Shengfeng Ye, Shiyu Wang, Shuiping Yu, Shunfeng Zhou, Shuting Pan, S.~S. Li, Shuang Zhou, Shaoqing Wu, Shengfeng Ye, Tao Yun, Tian Pei, Tianyu Sun, T.~Wang, Wangding Zeng, Wanjia Zhao, Wen Liu, Wenfeng Liang, Wenjun Gao, Wenqin Yu, Wentao Zhang, W.~L. Xiao, Wei An, Xiaodong Liu, Xiaohan Wang, Xiaokang Chen, Xiaotao Nie, Xin Cheng, Xin Liu, Xin Xie, Xingchao Liu, Xinyu Yang, Xinyuan Li, Xuecheng Su, Xuheng Lin, X.~Q. Li, Xiangyue Jin, Xiaojin Shen, Xiaosha Chen, Xiaowen Sun, Xiaoxiang Wang, Xinnan Song, Xinyi Zhou, Xianzu Wang, Xinxia Shan, Y.~K. Li, Y.~Q. Wang, Y.~X. Wei, Yang Zhang, Yanhong Xu, Yao Li, Yao Zhao, Yaofeng Sun, Yaohui Wang, Yi~Yu, Yichao Zhang, Yifan Shi, Yiliang Xiong, Ying He, Yishi Piao, Yisong Wang, Yixuan Tan, Yiyang Ma, Yiyuan Liu, Yongqiang Guo, Yuan Ou, Yuduan Wang, Yue Gong, Yuheng Zou, Yujia He, Yunfan Xiong, Yuxiang Luo, Yuxiang You, Yuxuan Liu, Yuyang Zhou, Y.~X. Zhu,
  Yanhong Xu, Yanping Huang, Yaohui Li, Yi~Zheng, Yuchen Zhu, Yunxian Ma, Ying Tang, Yukun Zha, Yuting Yan, Z.~Z. Ren, Zehui Ren, Zhangli Sha, Zhe Fu, Zhean Xu, Zhenda Xie, Zhengyan Zhang, Zhewen Hao, Zhicheng Ma, Zhigang Yan, Zhiyu Wu, Zihui Gu, Zijia Zhu, Zijun Liu, Zilin Li, Ziwei Xie, Ziyang Song, Zizheng Pan, Zhen Huang, Zhipeng Xu, Zhongyu Zhang, and Zhen Zhang.
\newblock Deepseek-r1: Incentivizing reasoning capability in llms via reinforcement learning, 2025.
\newblock URL \url{https://arxiv.org/abs/2501.12948}.

\bibitem[Gao et~al.(2024)Gao, Song, Yang, Cai, Miao, Dong, Li, Ma, Chen, Xu, Tang, Wang, Zan, Quan, Zhang, Sha, Zhang, Ren, Liu, and Chang]{omini-math}
Bofei Gao, Feifan Song, Zhe Yang, Zefan Cai, Yibo Miao, Qingxiu Dong, Lei Li, Chenghao Ma, Liang Chen, Runxin Xu, Zhengyang Tang, Benyou Wang, Daoguang Zan, Shanghaoran Quan, Ge~Zhang, Lei Sha, Yichang Zhang, Xuancheng Ren, Tianyu Liu, and Baobao Chang.
\newblock Omni-math: {A} universal olympiad level mathematic benchmark for large language models.
\newblock \emph{CoRR}, abs/2410.07985, 2024.
\newblock \doi{10.48550/ARXIV.2410.07985}.
\newblock URL \url{https://doi.org/10.48550/arXiv.2410.07985}.

\bibitem[Gou et~al.(2024)Gou, Shao, Gong, Shen, Yang, Duan, and Chen]{Critic}
Zhibin Gou, Zhihong Shao, Yeyun Gong, Yelong Shen, Yujiu Yang, Nan Duan, and Weizhu Chen.
\newblock {CRITIC:} large language models can self-correct with tool-interactive critiquing.
\newblock In \emph{The Twelfth International Conference on Learning Representations, {ICLR} 2024, Vienna, Austria, May 7-11, 2024}. OpenReview.net, 2024.
\newblock URL \url{https://openreview.net/forum?id=Sx038qxjek}.

\bibitem[Gui et~al.(2024)Gui, G{\^{a}}rbacea, and Veitch]{BoNBoN}
Lin Gui, Cristina G{\^{a}}rbacea, and Victor Veitch.
\newblock Bonbon alignment for large language models and the sweetness of best-of-n sampling.
\newblock \emph{CoRR}, abs/2406.00832, 2024.
\newblock \doi{10.48550/ARXIV.2406.00832}.
\newblock URL \url{https://doi.org/10.48550/arXiv.2406.00832}.

\bibitem[Hao et~al.(2023)Hao, Gu, Ma, Hong, Wang, Wang, and Hu]{RAP}
Shibo Hao, Yi~Gu, Haodi Ma, Joshua~Jiahua Hong, Zhen Wang, Daisy~Zhe Wang, and Zhiting Hu.
\newblock Reasoning with language model is planning with world model.
\newblock In Houda Bouamor, Juan Pino, and Kalika Bali (eds.), \emph{Proceedings of the 2023 Conference on Empirical Methods in Natural Language Processing, {EMNLP} 2023, Singapore, December 6-10, 2023}, pp.\  8154--8173. Association for Computational Linguistics, 2023.
\newblock \doi{10.18653/V1/2023.EMNLP-MAIN.507}.
\newblock URL \url{https://doi.org/10.18653/v1/2023.emnlp-main.507}.

\bibitem[Huang et~al.(2024{\natexlab{a}})Huang, Chen, Mishra, Zheng, Yu, Song, and Zhou]{LLM_cannot_self_correct}
Jie Huang, Xinyun Chen, Swaroop Mishra, Huaixiu~Steven Zheng, Adams~Wei Yu, Xinying Song, and Denny Zhou.
\newblock Large language models cannot self-correct reasoning yet.
\newblock In \emph{The Twelfth International Conference on Learning Representations, {ICLR} 2024, Vienna, Austria, May 7-11, 2024}. OpenReview.net, 2024{\natexlab{a}}.
\newblock URL \url{https://openreview.net/forum?id=IkmD3fKBPQ}.

\bibitem[Huang et~al.(2024{\natexlab{b}})Huang, Zou, Li, Liu, Zheng, Chern, Xia, Qin, Yuan, and Liu]{o1-journey2}
Zhen Huang, Haoyang Zou, Xuefeng Li, Yixiu Liu, Yuxiang Zheng, Ethan Chern, Shijie Xia, Yiwei Qin, Weizhe Yuan, and Pengfei Liu.
\newblock O1 replication journey -- part 2: Surpassing o1-preview through simple distillation, big progress or bitter lesson?, 2024{\natexlab{b}}.
\newblock URL \url{https://arxiv.org/abs/2411.16489}.

\bibitem[Jiang et~al.(2024)Jiang, Chen, Min, Chen, Cheng, Wang, Tang, Sun, Deng, Zhao, et~al.]{renda-1}
Jinhao Jiang, Zhipeng Chen, Yingqian Min, Jie Chen, Xiaoxue Cheng, Jiapeng Wang, Yiru Tang, Haoxiang Sun, Jia Deng, Wayne~Xin Zhao, et~al.
\newblock Technical report: Enhancing llm reasoning with reward-guided tree search.
\newblock \emph{arXiv preprint arXiv:2411.11694}, 2024.

\bibitem[Kamoi et~al.(2024)Kamoi, Zhang, Zhang, Han, and Zhang]{can-llm-self-revision}
Ryo Kamoi, Yusen Zhang, Nan Zhang, Jiawei Han, and Rui Zhang.
\newblock When can llms actually correct their own mistakes? {A} critical survey of self-correction of llms.
\newblock \emph{CoRR}, abs/2406.01297, 2024.
\newblock \doi{10.48550/ARXIV.2406.01297}.
\newblock URL \url{https://doi.org/10.48550/arXiv.2406.01297}.

\bibitem[Kool et~al.(2019)Kool, van Hoof, and Welling]{stochastic_beam_search}
Wouter Kool, Herke van Hoof, and Max Welling.
\newblock Stochastic beams and where to find them: The gumbel-top-k trick for sampling sequences without replacement.
\newblock In Kamalika Chaudhuri and Ruslan Salakhutdinov (eds.), \emph{Proceedings of the 36th International Conference on Machine Learning, {ICML} 2019, 9-15 June 2019, Long Beach, California, {USA}}, volume~97 of \emph{Proceedings of Machine Learning Research}, pp.\  3499--3508. {PMLR}, 2019.
\newblock URL \url{http://proceedings.mlr.press/v97/kool19a.html}.

\bibitem[Kumar et~al.(2024)Kumar, Zhuang, Agarwal, Su, Co{-}Reyes, Singh, Baumli, Iqbal, Bishop, Roelofs, Zhang, McKinney, Shrivastava, Paduraru, Tucker, Precup, Behbahani, and Faust]{SCoRE}
Aviral Kumar, Vincent Zhuang, Rishabh Agarwal, Yi~Su, John~D. Co{-}Reyes, Avi Singh, Kate Baumli, Shariq Iqbal, Colton Bishop, Rebecca Roelofs, Lei~M. Zhang, Kay McKinney, Disha Shrivastava, Cosmin Paduraru, George Tucker, Doina Precup, Feryal M.~P. Behbahani, and Aleksandra Faust.
\newblock Training language models to self-correct via reinforcement learning.
\newblock \emph{CoRR}, abs/2409.12917, 2024.
\newblock \doi{10.48550/ARXIV.2409.12917}.
\newblock URL \url{https://doi.org/10.48550/arXiv.2409.12917}.

\bibitem[Leike(2022)]{janleike2022why_excited_about_AI_assisted_human_feedback}
Jan Leike.
\newblock Why i’m excited about ai-assisted human feedback, 2022.
\newblock URL \url{https://substack.com/home/post/p-51216719}.

\bibitem[Lightman et~al.(2024)Lightman, Kosaraju, Burda, Edwards, Baker, Lee, Leike, Schulman, Sutskever, and Cobbe]{VerifySbyS}
Hunter Lightman, Vineet Kosaraju, Yuri Burda, Harrison Edwards, Bowen Baker, Teddy Lee, Jan Leike, John Schulman, Ilya Sutskever, and Karl Cobbe.
\newblock Let's verify step by step.
\newblock In \emph{The Twelfth International Conference on Learning Representations, {ICLR} 2024, Vienna, Austria, May 7-11, 2024}. OpenReview.net, 2024.
\newblock URL \url{https://openreview.net/forum?id=v8L0pN6EOi}.

\bibitem[Luo et~al.(2025)Luo, Shen, He, Wang, Liu, Li, Tan, Cao, and Tao]{o1-pruner}
Haotian Luo, Li~Shen, Haiying He, Yibo Wang, Shiwei Liu, Wei Li, Naiqiang Tan, Xiaochun Cao, and Dacheng Tao.
\newblock O1-pruner: Length-harmonizing fine-tuning for o1-like reasoning pruning.
\newblock \emph{ArXiv}, abs/2501.12570, 2025.
\newblock URL \url{https://api.semanticscholar.org/CorpusID:275790112}.

\bibitem[Madaan et~al.(2023)Madaan, Tandon, Gupta, Hallinan, Gao, Wiegreffe, Alon, Dziri, Prabhumoye, Yang, Gupta, Majumder, Hermann, Welleck, Yazdanbakhsh, and Clark]{Self_refine}
Aman Madaan, Niket Tandon, Prakhar Gupta, Skyler Hallinan, Luyu Gao, Sarah Wiegreffe, Uri Alon, Nouha Dziri, Shrimai Prabhumoye, Yiming Yang, Shashank Gupta, Bodhisattwa~Prasad Majumder, Katherine Hermann, Sean Welleck, Amir Yazdanbakhsh, and Peter Clark.
\newblock Self-refine: Iterative refinement with self-feedback.
\newblock In Alice Oh, Tristan Naumann, Amir Globerson, Kate Saenko, Moritz Hardt, and Sergey Levine (eds.), \emph{Advances in Neural Information Processing Systems 36: Annual Conference on Neural Information Processing Systems 2023, NeurIPS 2023, New Orleans, LA, USA, December 10 - 16, 2023}, 2023.
\newblock URL \url{http://papers.nips.cc/paper\_files/paper/2023/hash/91edff07232fb1b55a505a9e9f6c0ff3-Abstract-Conference.html}.

\bibitem[Min et~al.(2024)Min, Chen, Jiang, Chen, Deng, Hu, Tang, Wang, Cheng, Song, et~al.]{renda-2}
Yingqian Min, Zhipeng Chen, Jinhao Jiang, Jie Chen, Jia Deng, Yiwen Hu, Yiru Tang, Jiapeng Wang, Xiaoxue Cheng, Huatong Song, et~al.
\newblock Imitate, explore, and self-improve: A reproduction report on slow-thinking reasoning systems.
\newblock \emph{arXiv preprint arXiv:2412.09413}, 2024.

\bibitem[Muennighoff et~al.(2025)Muennighoff, Yang, Shi, Li, Fei-Fei, Hajishirzi, Zettlemoyer, Liang, Candès, and Hashimoto]{s1}
Niklas Muennighoff, Zitong Yang, Weijia Shi, Xiang~Lisa Li, Li~Fei-Fei, Hannaneh Hajishirzi, Luke Zettlemoyer, Percy Liang, Emmanuel Candès, and Tatsunori Hashimoto.
\newblock s1: Simple test-time scaling, 2025.
\newblock URL \url{https://arxiv.org/abs/2501.19393}.

\bibitem[OpenAI(2024{\natexlab{a}})]{o1_blog}
OpenAI.
\newblock Learning to reason with llms, 2024{\natexlab{a}}.
\newblock URL \url{https://openai.com/index/learning-to-reason-with-llms/}.

\bibitem[OpenAI(2024{\natexlab{b}})]{o1_system_card}
OpenAI.
\newblock Openai o1 system card, 2024{\natexlab{b}}.
\newblock URL \url{https://cdn.openai.com/o1-system-card-20241205.pdf}.

\bibitem[Qin et~al.(2024)Qin, Li, Zou, Liu, Xia, Huang, Ye, Yuan, Liu, Li, and Liu]{o1-journey1}
Yiwei Qin, Xuefeng Li, Haoyang Zou, Yixiu Liu, Shijie Xia, Zhen Huang, Yixin Ye, Weizhe Yuan, Hector Liu, Yuanzhi Li, and Pengfei Liu.
\newblock {O1} replication journey: {A} strategic progress report - part 1.
\newblock \emph{CoRR}, abs/2410.18982, 2024.
\newblock \doi{10.48550/ARXIV.2410.18982}.
\newblock URL \url{https://doi.org/10.48550/arXiv.2410.18982}.

\bibitem[Qiu et~al.(2024)Qiu, Lu, Zeng, Guo, Geng, Wang, Huang, Wu, and Wang]{TreeBoN}
Jiahao Qiu, Yifu Lu, Yifan Zeng, Jiacheng Guo, Jiayi Geng, Huazheng Wang, Kaixuan Huang, Yue Wu, and Mengdi Wang.
\newblock Treebon: Enhancing inference-time alignment with speculative tree-search and best-of-n sampling.
\newblock \emph{CoRR}, abs/2410.16033, 2024.
\newblock \doi{10.48550/ARXIV.2410.16033}.
\newblock URL \url{https://doi.org/10.48550/arXiv.2410.16033}.

\bibitem[Rein et~al.(2023)Rein, Hou, Stickland, Petty, Pang, Dirani, Michael, and Bowman]{gpqa}
David Rein, Betty~Li Hou, Asa~Cooper Stickland, Jackson Petty, Richard~Yuanzhe Pang, Julien Dirani, Julian Michael, and Samuel~R. Bowman.
\newblock {GPQA:} {A} graduate-level google-proof q{\&}a benchmark.
\newblock \emph{CoRR}, abs/2311.12022, 2023.
\newblock \doi{10.48550/ARXIV.2311.12022}.
\newblock URL \url{https://doi.org/10.48550/arXiv.2311.12022}.

\bibitem[Sessa et~al.(2024)Sessa, Dadashi, Hussenot, Ferret, Vieillard, Ram{\'{e}}, Shahriari, Perrin, Friesen, Cideron, Girgin, Stanczyk, Michi, Sinopalnikov, Ramos, H{\'{e}}liou, Severyn, Hoffman, Momchev, and Bachem]{BOND}
Pier~Giuseppe Sessa, Robert Dadashi, L{\'{e}}onard Hussenot, Johan Ferret, Nino Vieillard, Alexandre Ram{\'{e}}, Bobak Shahriari, Sarah Perrin, Abe Friesen, Geoffrey Cideron, Sertan Girgin, Piotr Stanczyk, Andrea Michi, Danila Sinopalnikov, Sabela Ramos, Am{\'{e}}lie H{\'{e}}liou, Aliaksei Severyn, Matt Hoffman, Nikola Momchev, and Olivier Bachem.
\newblock {BOND:} aligning llms with best-of-n distillation.
\newblock \emph{CoRR}, abs/2407.14622, 2024.
\newblock \doi{10.48550/ARXIV.2407.14622}.
\newblock URL \url{https://doi.org/10.48550/arXiv.2407.14622}.

\bibitem[Snell et~al.(2024)Snell, Lee, Xu, and Kumar]{Scaling_test_time_compute}
Charlie Snell, Jaehoon Lee, Kelvin Xu, and Aviral Kumar.
\newblock Scaling {LLM} test-time compute optimally can be more effective than scaling model parameters.
\newblock \emph{CoRR}, abs/2408.03314, 2024.
\newblock \doi{10.48550/ARXIV.2408.03314}.
\newblock URL \url{https://doi.org/10.48550/arXiv.2408.03314}.

\bibitem[Sun et~al.(2024)Sun, Haider, Zhang, Yang, Qiu, Yin, Wang, Bartlett, and Zanette]{Speculative_bon}
Hanshi Sun, Momin Haider, Ruiqi Zhang, Huitao Yang, Jiahao Qiu, Ming Yin, Mengdi Wang, Peter~L. Bartlett, and Andrea Zanette.
\newblock Fast best-of-n decoding via speculative rejection.
\newblock \emph{CoRR}, abs/2410.20290, 2024.
\newblock \doi{10.48550/ARXIV.2410.20290}.
\newblock URL \url{https://doi.org/10.48550/arXiv.2410.20290}.

\bibitem[Team(2024)]{qwq}
Qwen Team.
\newblock Qwq: Reflect deeply on the boundaries of the unknown, November 2024.
\newblock URL \url{https://qwenlm.github.io/blog/qwq-32b-preview/}.

\bibitem[Wan et~al.(2024)Wan, Feng, Wen, McAleer, Wen, Zhang, and Wang]{Alpha-zero-like}
Ziyu Wan, Xidong Feng, Muning Wen, Stephen~Marcus McAleer, Ying Wen, Weinan Zhang, and Jun Wang.
\newblock Alphazero-like tree-search can guide large language model decoding and training.
\newblock In \emph{Forty-first International Conference on Machine Learning, {ICML} 2024, Vienna, Austria, July 21-27, 2024}. OpenReview.net, 2024.
\newblock URL \url{https://openreview.net/forum?id=C4OpREezgj}.

\bibitem[Wang et~al.(2023)Wang, Wei, Schuurmans, Le, Chi, Narang, Chowdhery, and Zhou]{SC}
Xuezhi Wang, Jason Wei, Dale Schuurmans, Quoc~V. Le, Ed~H. Chi, Sharan Narang, Aakanksha Chowdhery, and Denny Zhou.
\newblock Self-consistency improves chain of thought reasoning in language models.
\newblock In \emph{The Eleventh International Conference on Learning Representations, {ICLR} 2023, Kigali, Rwanda, May 1-5, 2023}. OpenReview.net, 2023.
\newblock URL \url{https://openreview.net/forum?id=1PL1NIMMrw}.

\bibitem[Wang et~al.(2025)Wang, Liu, Xu, Liang, Chen, He, Song, Yu, Li, Zhang, Wang, Tu, Mi, and Yu]{o1-underthink}
Yue Wang, Qiuzhi Liu, Jiahao Xu, Tian Liang, Xingyu Chen, Zhiwei He, Linfeng Song, Dian Yu, Juntao Li, Zhuosheng Zhang, Rui Wang, Zhaopeng Tu, Haitao Mi, and Dong Yu.
\newblock Thoughts are all over the place: On the underthinking of o1-like llms, 2025.
\newblock URL \url{https://arxiv.org/abs/2501.18585}.

\bibitem[Wei et~al.(2022)Wei, Wang, Schuurmans, Bosma, Ichter, Xia, Chi, Le, and Zhou]{wei2022chain}
Jason Wei, Xuezhi Wang, Dale Schuurmans, Maarten Bosma, Brian Ichter, Fei Xia, Ed~H. Chi, Quoc~V. Le, and Denny Zhou.
\newblock Chain-of-thought prompting elicits reasoning in large language models.
\newblock In Sanmi Koyejo, S.~Mohamed, A.~Agarwal, Danielle Belgrave, K.~Cho, and A.~Oh (eds.), \emph{Advances in Neural Information Processing Systems 35: Annual Conference on Neural Information Processing Systems 2022, NeurIPS 2022, New Orleans, LA, USA, November 28 - December 9, 2022}, 2022.
\newblock URL \url{http://papers.nips.cc/paper\_files/paper/2022/hash/9d5609613524ecf4f15af0f7b31abca4-Abstract-Conference.html}.

\bibitem[Xie et~al.(2023)Xie, Kawaguchi, Zhao, Zhao, Kan, He, and Xie]{Self_evaluation_guided_beam_search}
Yuxi Xie, Kenji Kawaguchi, Yiran Zhao, James~Xu Zhao, Min{-}Yen Kan, Junxian He, and Michael~Qizhe Xie.
\newblock Self-evaluation guided beam search for reasoning.
\newblock In Alice Oh, Tristan Naumann, Amir Globerson, Kate Saenko, Moritz Hardt, and Sergey Levine (eds.), \emph{Advances in Neural Information Processing Systems 36: Annual Conference on Neural Information Processing Systems 2023, NeurIPS 2023, New Orleans, LA, USA, December 10 - 16, 2023}, 2023.
\newblock URL \url{http://papers.nips.cc/paper\_files/paper/2023/hash/81fde95c4dc79188a69ce5b24d63010b-Abstract-Conference.html}.

\bibitem[Yang et~al.(2024)Yang, Zhang, Hui, Gao, Yu, Li, Liu, Tu, Zhou, Lin, Lu, Xue, Lin, Liu, Ren, and Zhang]{Qwen2.5-Math}
An~Yang, Beichen Zhang, Binyuan Hui, Bofei Gao, Bowen Yu, Chengpeng Li, Dayiheng Liu, Jianhong Tu, Jingren Zhou, Junyang Lin, Keming Lu, Mingfeng Xue, Runji Lin, Tianyu Liu, Xingzhang Ren, and Zhenru Zhang.
\newblock Qwen2.5-math technical report: Toward mathematical expert model via self-improvement.
\newblock \emph{CoRR}, abs/2409.12122, 2024.
\newblock \doi{10.48550/ARXIV.2409.12122}.
\newblock URL \url{https://doi.org/10.48550/arXiv.2409.12122}.

\bibitem[Ye et~al.(2025)Ye, Huang, Xiao, Chern, Xia, and Liu]{ye2025limoreasoning}
Yixin Ye, Zhen Huang, Yang Xiao, Ethan Chern, Shijie Xia, and Pengfei Liu.
\newblock Limo: Less is more for reasoning, 2025.
\newblock URL \url{https://arxiv.org/abs/2502.03387}.

\bibitem[Yu et~al.(2024)Yu, Gao, and Wang]{OVM}
Fei Yu, Anningzhe Gao, and Benyou Wang.
\newblock Ovm, outcome-supervised value models for planning in mathematical reasoning.
\newblock In Kevin Duh, Helena G{\'{o}}mez{-}Adorno, and Steven Bethard (eds.), \emph{Findings of the Association for Computational Linguistics: {NAACL} 2024, Mexico City, Mexico, June 16-21, 2024}, pp.\  858--875. Association for Computational Linguistics, 2024.
\newblock \doi{10.18653/V1/2024.FINDINGS-NAACL.55}.
\newblock URL \url{https://doi.org/10.18653/v1/2024.findings-naacl.55}.

\bibitem[Zeng et~al.(2024)Zeng, Cheng, Yin, Wang, Li, Zhou, Guo, Huang, and Qiu]{o1-roadmap}
Zhiyuan Zeng, Qinyuan Cheng, Zhangyue Yin, Bo~Wang, Shimin Li, Yunhua Zhou, Qipeng Guo, Xuanjing Huang, and Xipeng Qiu.
\newblock Scaling of search and learning: {A} roadmap to reproduce o1 from reinforcement learning perspective.
\newblock \emph{CoRR}, abs/2412.14135, 2024.
\newblock \doi{10.48550/ARXIV.2412.14135}.
\newblock URL \url{https://doi.org/10.48550/arXiv.2412.14135}.

\bibitem[Zhang et~al.(2023)Zhang, Chen, Shen, Ding, Tenenbaum, and Gan]{MCTS_for_code_generation}
Shun Zhang, Zhenfang Chen, Yikang Shen, Mingyu Ding, Joshua~B. Tenenbaum, and Chuang Gan.
\newblock Planning with large language models for code generation.
\newblock In \emph{The Eleventh International Conference on Learning Representations, {ICLR} 2023, Kigali, Rwanda, May 1-5, 2023}. OpenReview.net, 2023.
\newblock URL \url{https://openreview.net/forum?id=Lr8cOOtYbfL}.

\bibitem[Zhang et~al.(2024)Zhang, Khalifa, Logeswaran, Kim, Lee, Lee, and Wang]{teach-self-revision}
Yunxiang Zhang, Muhammad Khalifa, Lajanugen Logeswaran, Jaekyeom Kim, Moontae Lee, Honglak Lee, and Lu~Wang.
\newblock Small language models need strong verifiers to self-correct reasoning.
\newblock In Lun{-}Wei Ku, Andre Martins, and Vivek Srikumar (eds.), \emph{Findings of the Association for Computational Linguistics, {ACL} 2024, Bangkok, Thailand and virtual meeting, August 11-16, 2024}, pp.\  15637--15653. Association for Computational Linguistics, 2024.
\newblock \doi{10.18653/V1/2024.FINDINGS-ACL.924}.
\newblock URL \url{https://doi.org/10.18653/v1/2024.findings-acl.924}.

\bibitem[Zheng et~al.(2024)Zheng, Yin, Xie, Sun, Huang, Yu, Cao, Kozyrakis, Stoica, Gonzalez, Barrett, and Sheng]{sglang}
Lianmin Zheng, Liangsheng Yin, Zhiqiang Xie, Chuyue Sun, Jeff Huang, Cody~Hao Yu, Shiyi Cao, Christos Kozyrakis, Ion Stoica, Joseph~E. Gonzalez, Clark~W. Barrett, and Ying Sheng.
\newblock Sglang: Efficient execution of structured language model programs.
\newblock In Amir Globersons, Lester Mackey, Danielle Belgrave, Angela Fan, Ulrich Paquet, Jakub~M. Tomczak, and Cheng Zhang (eds.), \emph{Advances in Neural Information Processing Systems 38: Annual Conference on Neural Information Processing Systems 2024, NeurIPS 2024, Vancouver, BC, Canada, December 10 - 15, 2024}, 2024.
\newblock URL \url{http://papers.nips.cc/paper\_files/paper/2024/hash/724be4472168f31ba1c9ac630f15dec8-Abstract-Conference.html}.

\end{thebibliography}
\bibliographystyle{colm2024_conference}
\clearpage

\appendix
\section{Is Invalid Scaling Phenomenon Conflict to Findings of R1 technique Report?}





The training objective of R1 aims to improve model accuracy, yet we observe that correct solutions tend to be shorter than incorrect ones. This raises an intriguing question: Why does R1's reinforcement learning (RL) training consistently produce longer solutions?

\begin{figure}
    \centering
    \includegraphics[width=1.0\linewidth]{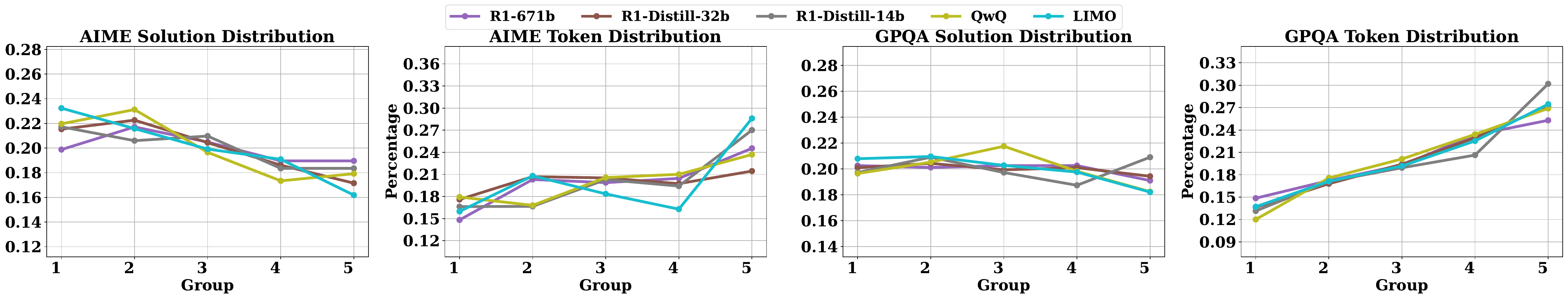}
    \caption{The number of correct solutions and tokens distributed across groups of different lengths.}
    \label{fig:overall-distribution}
\end{figure}

To investigate this phenomenon, we analyzed five solutions per question, organizing them into groups by length in ascending order. Figure \ref{fig:overall-distribution} illustrates the distribution of correct solutions across these groups.

Our analysis revealed that correct solutions predominantly appear in shorter-length groups, particularly in the AIME dataset. However, when examining the token distribution, we found that correct solution tokens are concentrated in longer-solution groups. This apparent contradiction arises because the total token count is determined by both the number of solutions and the average tokens per solution. As shown in Figure \ref{fig:overall-len}, solutions in the longest group contain nearly twice as many tokens as those in the shortest group. This explains why, despite having fewer individual solutions, longer solutions account for a greater share of the total tokens.


We hypothesize that this discrepancy explains why RL training tends to produce longer solutions: the training process may favor generating longer solutions, even if they are less accurate, because they contribute more tokens to the gradient. 

\begin{figure*}[t]
    \centering
    \begin{subfigure}[b]{0.3\textwidth}
        \includegraphics[width=\textwidth]{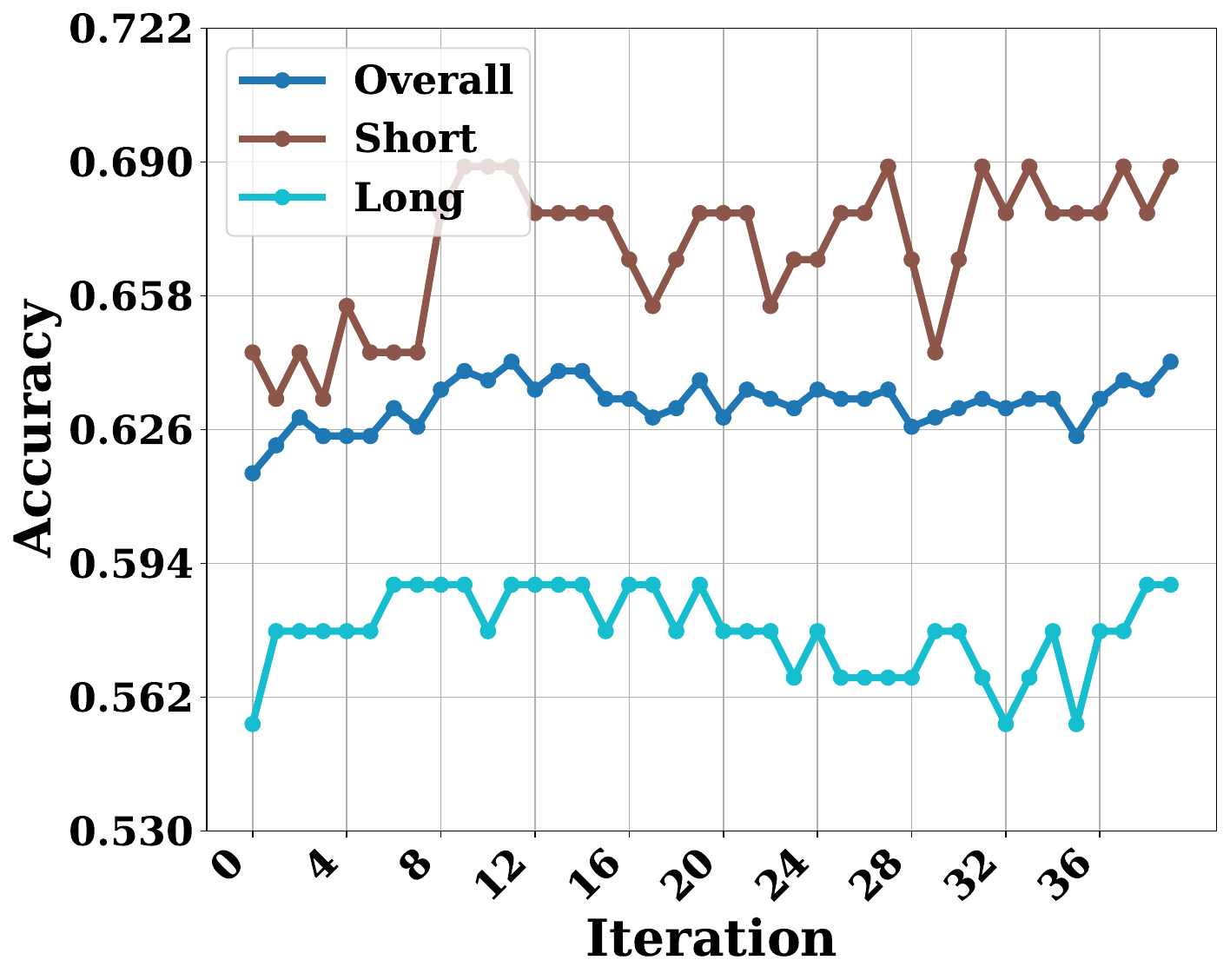}
        \caption{R1-Distill-32b}
    \end{subfigure}
    \begin{subfigure}[b]{0.3\textwidth}
        \includegraphics[width=\textwidth]{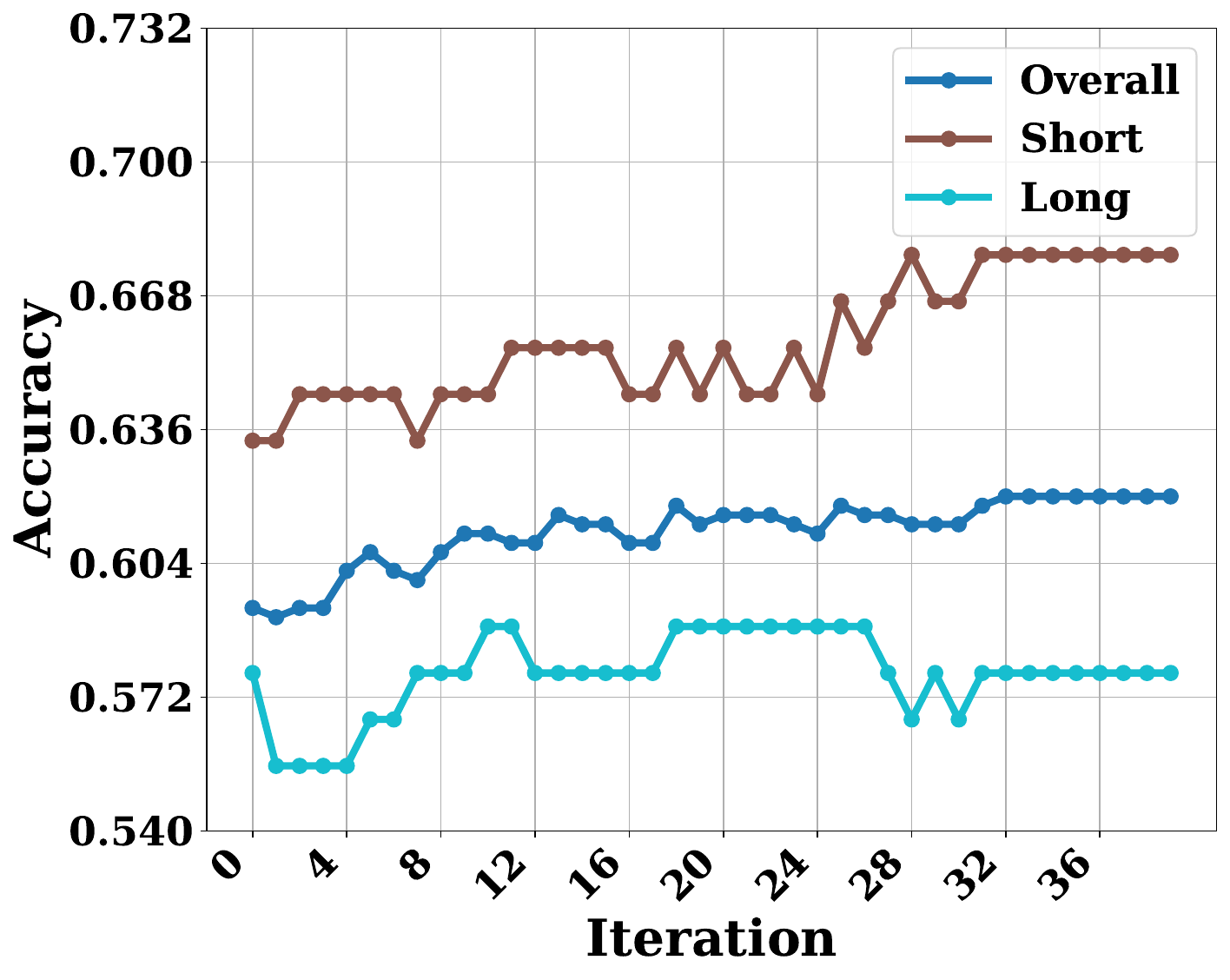}
        \caption{R1-Distill-14b}
    \end{subfigure}
    \begin{subfigure}[b]{0.3\textwidth}
        \includegraphics[width=\textwidth]{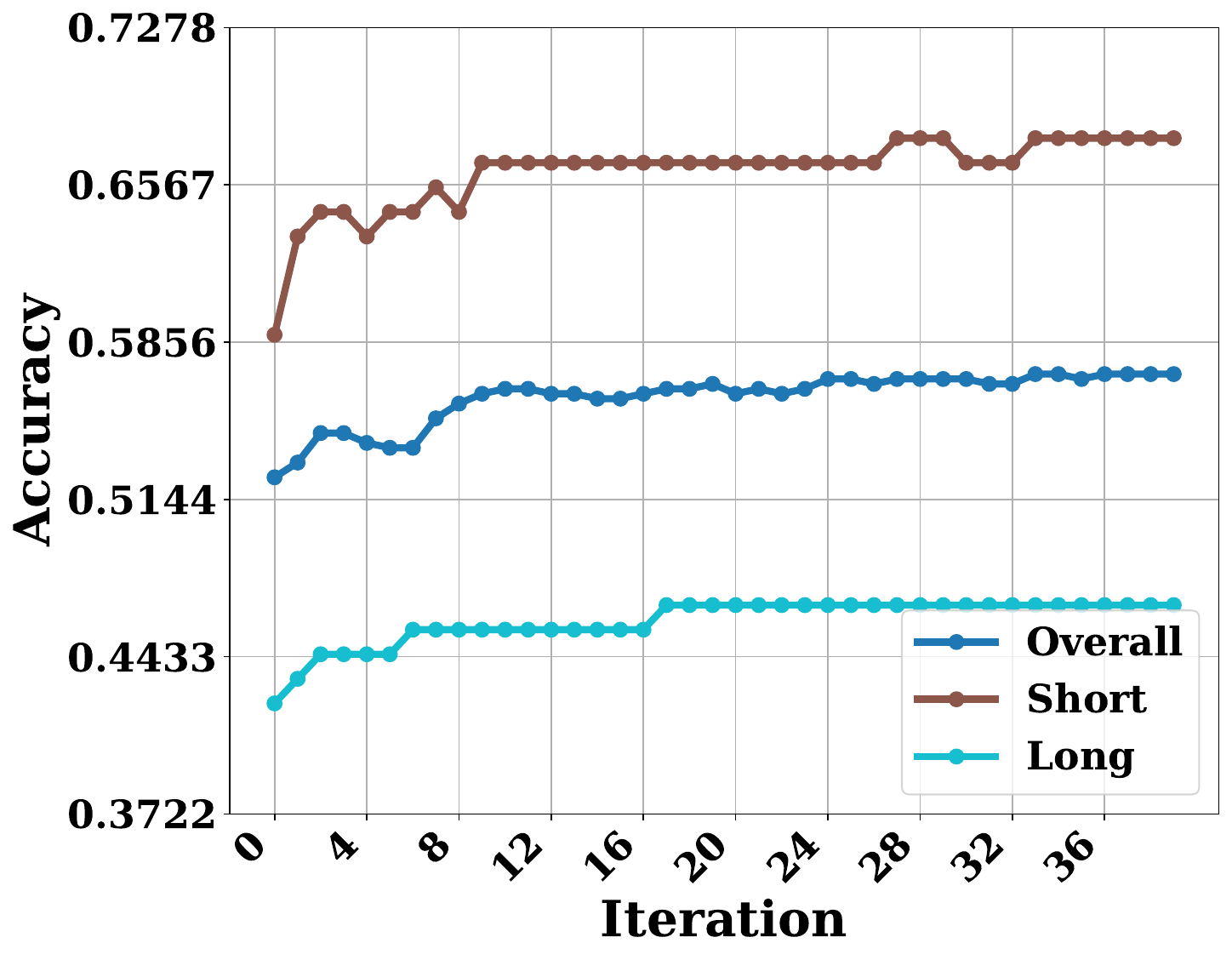}
        \caption{R1-Distill-14b}
    \end{subfigure}
    \caption{Accuracy of short solutions and long solutions of R1-Distill-14b (a) and R1-Distill-32b (b) during sequential revision.}
    \label{fig:short-long-revision}
\end{figure*}

\begin{figure*}[t]
    \centering
    \includegraphics[width=\textwidth]{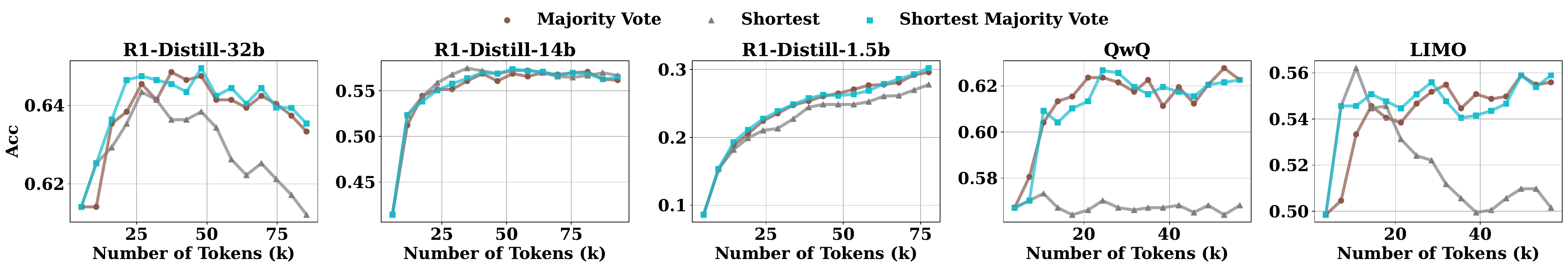}
    \caption{Performance Comparison between Majority Vote and Shortest Majority Vote on GPQA.}
    \label{fig:short-majority-vote-gpqa}
\end{figure*}

\section{Further analysis on Sequential Scaling on R1-Distill-14b, R1-Distill-32b and LIMO}\label{app:short-long-revision}
In Section \ref{sec:sequential-revision}, we observed that R1-Distill-14b, R1-Distill-32b and LIMO demonstrated some performance improvements after multiple rounds of self-revision, followed by stabilization. Furthermore, in Section \ref{sec:invalid-scaling}, we found that the correct solutions generated by R1-Distill-14b, R1-Distill-32b and LIMO were generally shorter than incorrect solutions. To reconcile these seemingly contradictory findings and further analyze how R1-Distill-14b, R1-Distill-32b and LIMO benefit from self-revision, we conducted a detailed analysis of self-revision outcomes on both long and short solutions. Our methodology for collecting long and short solutions involved sampling five solutions for each question, ordering them by length, and then segregating the longest and shortest solutions into separate groups. The results of self-revision on both short and long solutions are presented in Figure \ref{fig:short-long-revision}. Our analysis reveals that short solutions exhibited significant performance improvements following self-revision, while this trend was less pronounced for long solutions. Therefore, the performance improvements we observed through self-revision in R1-Distill-14b, R1-Distill-32b and LIMO primarily stem from the self-revision on short solutions. This suggests that the relationship between accuracy and solution length for these models is complex, demonstrating neither a strictly positive nor negative correlation with length.

\section{Parallel Scaling of Shortest Majority Vote on GPQA}\label{app:gpqa-scale}
In Section \ref{sec:new-method}, we demonstrated that our proposed Shortest Majority Vote achieves superior test-time scaling performance compared to the other two methods on the AIME benchmark. In this section, we present the parallel-scaling results on GPQA in Figure \ref{fig:short-majority-vote-gpqa}. While Shortest Majority Vote consistently outperforms the Shortest method on GPQA, it does not exhibit significantly better parallel scaling performance compared to Majority Vote on this benchmark. This phenomenon might be attributed to the smaller performance gap between short and long solutions on GPQA compared to AIME, suggesting that solution length plays a less critical role in determining solution quality on the GPQA benchmark, which can be observed from Figure \ref{fig:overall-acc}

\section{Prompt}\label{app:prompt}

System prompt:
\begin{promptbox}{System prompt}
You are a helpful and harmless assistant. You should think step-by-step.
\end{promptbox}

Instruction for MATH-500, AIME and Omini-MATH:
\begin{promptbox}{Instruction}
Answer the question and enclose the final answer in boxed\{\}
\end{promptbox}

Instruction for GPQA:
\begin{promptbox}{Instruction}
Select the best answer from the following options. Output only the letter corresponding to the correct answer, enclosed in boxed\{\}.
\end{promptbox}

\section{Examples of self-revision}\label{app:revision-examples}
\begin{promptbox}{Examples}\label{fig:revision-example}
Wait, let me verify that again ...\\\\
Wait, but that seems straightforward, but let me check if I got the constants right ...\\\\
Wait, but let me verify this to ensure I didn't make a mistake ...\\\\
Wait, so is the answer 756? But let me check if this is consistent ...\\\\
Wait, but in 3D space, the centers might not be coplanar? ...\\\\
Alternatively, try to find a general formula ...\\\\
Alternatively, consider that m is such that m divides k where k is from 1 to 999 ...\\\\
Alternatively, maybe we can use modulo 8 to get constraints ...\\\\
Alternatively, perhaps there's a smarter approach ...\\\\
Alternatively, another way to think about this problem is to recognize that w and z are roots of unity ...
\end{promptbox}

\newpage

\end{document}